\documentclass[11pt]{article}

\usepackage{microtype}
\usepackage{graphicx}
\usepackage{subcaption}
\usepackage{booktabs} %
\usepackage[top=1in,bottom=1in,left=1in,right=1in]{geometry}
\usepackage{natbib}
\usepackage{algorithm}
\usepackage{algorithmic}

\usepackage{hyperref}
\makeatletter
\let\Hy@linktoc\Hy@linktoc@none
\makeatother

\usepackage{xspace}
\usepackage{bbm}
\usepackage[shortlabels]{enumitem}
\usepackage{multirow}
\usepackage{makecell}

\newcommand{\FL}{FL\xspace}

\newcommand{\DDP}{distributed DP\xspace}

\newcommand{\SA}{SecAgg\xspace}
\newcommand{\puc}{privacy-utility-communication\xspace}

\usepackage{fixltx2e}
\usepackage{fancyvrb}
\RecustomVerbatimCommand{\VerbatimInput}{VerbatimInput}%
{fontsize=\footnotesize,
 framesep=2em, %
 rulecolor=\color{gray},
 commandchars=\|\(\), %
 commentchar=*        %
}
\newcommand{\lp}{\left(}
\newcommand{\rp}{\right)}
\newcommand{\lb}{\left[}
\newcommand{\rb}{\right]}
\newcommand{\lbp}{\left\{}
\newcommand{\rbp}{\right\}}
\newcommand{\lba}{\left\lvert}
\newcommand{\rba}{\right\rvert}
\newcommand{\lV}{\left\lVert}
\newcommand{\rV}{\right\rVert}
\newcommand{\mv}{\middle\vert}

\newcommand{\mcal}{\mathcal}

\newcommand{\mb}{\mathbf}
\newcommand{\bbm}{\mathbbm}
\newcommand{\mbb}{\mathbb}
\newcommand{\msf}{\mathsf}

\newcommand{\ra}{\rightarrow}

\newcommand{\lan}{\langle}
\newcommand{\ran}{\rangle}

\newcommand{\eqDef}{\triangleq}
\newcommand{\diid}{\overset{\text{i.i.d.}}{\sim}}

\newcommand{\E}{\mathbb{E}}

\renewcommand{\Pr}{\mathbb{P}}

\usepackage{amsmath}
\usepackage{amssymb}
\usepackage{mathtools}
\usepackage{amsthm}

\usepackage[capitalize,noabbrev]{cleveref}

\theoremstyle{plain}
\newtheorem{theorem}{Theorem}[section]

\newtheorem{lemma}[theorem]{Lemma}
\newtheorem{corollary}[theorem]{Corollary}
\theoremstyle{definition}
\newtheorem{definition}[theorem]{Definition}

\theoremstyle{remark}
\newtheorem{remark}[theorem]{Remark}

\usepackage[textsize=tiny]{todonotes}

\title{\vspace{-1em} \textbf{The Fundamental Price of Secure Aggregation in Differentially Private Federated Learning}  \vspace{1ex} } 
\author{ Wei-Ning Chen$^{* \dag \ddag}$ \and Christopher A. Choquette-Choo$^{* \ddag}$ \and Peter Kairouz$^{* \ddag}$ \and Ananda Theertha Suresh$^{ * \ddag}$ } 
\date{%
Stanford University$^{ \dag}$, Google Research$^{ \ddag}$\\[2ex]%
\texttt{wnchen@stanford.edu, \{cchoquette, kairouz, theertha\}@google.com}
}

\begin{document}

\maketitle 

\begin{abstract}
We consider the problem of training a $d$ dimensional model with distributed differential privacy (DP) where secure aggregation (SecAgg) is used to ensure that the server only sees the noisy sum of $n$ model updates in every training round. Taking into account the constraints imposed by SecAgg, we characterize the fundamental communication cost required to obtain the best accuracy achievable under $\varepsilon$ central DP (i.e. under a fully trusted server and no communication constraints). Our results show that $\tilde{O}\lp \min(n^2\varepsilon^2, d) \rp$ bits per client are both sufficient and necessary, and this fundamental limit can be achieved by a linear scheme based on sparse random projections. This provides a significant improvement relative to state-of-the-art SecAgg distributed DP schemes which use $\tilde{O}(d\log(d/\varepsilon^2))$ bits per client. 
  
Empirically, we evaluate our proposed scheme on real-world federated learning tasks. We find that our theoretical analysis is well matched in practice. In particular, we show that we can reduce the communication cost significantly to under $1.2$ bits per parameter in realistic privacy settings without decreasing test-time performance. Our work hence theoretically and empirically specifies the fundamental price of using SecAgg. \let\thefootnote\relax\footnotetext{* Equal contribution with alphabetical authorship. \dag Work completed while on internship at Google.}
\end{abstract}

\newpage
\tableofcontents

\newpage

\section{Introduction}\label{sec:intro}
Federated learning (FL) is a widely used machine learning framework where multiple clients collaborate in learning a model under the coordination of a central server~\citep{mcmahan2017communication,aopfl}. 
One of the primary attractions of FL is that it provides data confidentiality and can provide a level of privacy to participating clients through data minimization: the raw client data never leaves the device, and only updates to models (e.g., gradient updates) are sent back to the central server. 
This provides practical privacy improvements over centralized settings because updates typically contain less information about the clients, because they are more focused on the learning task, and also only need to be held ephemerally by the server

However, this vanilla federated learning does not provide any formal or provable privacy guarantees.
To do so, FL is often combined with  differential privacy (DP)~\citep{dwork2006calibrating}. 
This can be done in one of two ways\footnote{Local DP is yet another alternative, but it incurs higher utility loss and is therefore not typically used in practice.}: 1) perturbing the aggregated (local) model updates at the server before updating the global model, or 2) perturbing each client's model update locally and using a cryptographic multi-party computation protocol to ensure that the server only sees the noisy aggregate.
The former is referred to as \emph{central} DP, and it relies on the clients' trust in the server because any sensitive information contained in the model updates is revealed to and temporally stored on the server. The latter is referred to as \emph{distributed} DP \citep{dwork2006our, kairouz2021distributed, agarwal2021skellam, agarwal2018cpsgd}, and it offers privacy guarantees with respect to an \emph{honest-but-curious} server. Thus, a key technology for formalizing and strengthening FL’s privacy guarantees is a secure vector sum protocol called secure aggregation (SecAgg)~\citep{bonawitz2016practical, bell2020secure}, which lets the server see the aggregate client updates but not the individual ones.

Despite enhancing the clients' privacy, aggregating model updates via SecAgg drastically increases the computation and communication overheads~\citep{bonawitz2016practical, bonawitz2019towards}. This is even worse in federated settings where communication occurs over bandwidth-limited wireless links, and the extra communication costs may become a bottleneck that hampers efficient training of large-scale machine learning models. For example,~\citet{kairouz2021distributed} reports that when training a language model with SecAgg and DP, even with a carefully designed quantization scheme, each client still needs to transmit about $16$ bits \emph{per model parameter} each round. Moreover, the bitwidth needs to be scaled up when the privacy requirements are more stringent. This behavior disobeys the conclusion of~\citet{chen2020breaking} (derived under local DP), which shows that the optimal communication cost should \emph{decay} with the privacy budget, i.e., data is more compressible in the high privacy regime.

Furthermore, because the server aggregates the model updates via SecAgg, we can only compress the model updates locally using linear schemes. This constraint rules out many popular compression schemes such as entropy encoders or gradient sparsification \citep{aji2017sparse, lin2017deep, wangni2017gradient, havasi2018minimal, oktay2019scalable} etc., as these methods are non-linear. 

Therefore, it is unclear whether or not the communication cost of SecAgg reported in~\citet{kairouz2021distributed, agarwal2021skellam, agarwal2018cpsgd} is fundamental. If not, what is the smallest communication needed to achieve distributed DP with secure aggregation to achieve the same performance as in the centralized DP setting?

In this paper, we answer the above question, showing that the communication costs of existing mechanisms are strictly sub-optimal in the distributed mean estimation (DME) task \citep{an2016distributed} (see Section~\ref{sec:formulation} for details). We also propose a SecAgg comaptible \emph{linear} compression scheme based on sparse random projections (Algorithm~\ref{alg:sketch_mean_decoder}), and then combine it with the distributed discrete Gaussian (DDG) mechanism proposed by~\citet{kairouz2021distributed}. Theoretically, we prove that our scheme requires $\tilde{O}\lp \min(n^2\varepsilon^2, d) \rp$ bits per client, where $n$ is the per-round number of clients. This cost is significantly smaller than the communication cost of previous schemes which was $\tilde{O}_\delta(d\log(d/\varepsilon^2))$ bits per client\footnote{For simplicity, we use the $\tilde{O}_\delta\lp \cdot\rp$ notation to hide the dependency on $\delta$ and $\log n$}. To give perspective, 
(1) $n$ is usually on the order of $10^3$ per round due to the computational overhead of SecAgg, and (2) $\varepsilon$ is the privacy budget for a single round, i.e., $\varepsilon \approx {\varepsilon_{\msf{final}}}/{\sqrt{R}}$ if there are $R$ training rounds. Thus, for practical FL settings where large models are trained with SecAgg over many rounds, $n^2\varepsilon^2$ is typically (much) smaller than $d$. 

We complement our achievability results with a matching lower bound, showing that to obtain an unbiased estimator of the mean vector, each client needs to communicate $\tilde{\Omega}(\min(n^2\varepsilon^2, d))$ bits with the server. Our upper and lower bounds together specify the fundamental privacy-communication-accuracy trade-offs under SecAgg and DP.

In addition, we show that with additional sparsity assumptions, we can further improve both the accuracy and communication efficiency while achieving the same privacy requirement, leading to a logarithmic dependency on $d$.

Empirically, we verify our scheme on a variety of real-world FL tasks. Compared to existing distributed DP schemes, we observe $10$x or more compression with no significant decrease in test-time performance. Moreover, the compression rates can be made even higher with tighter privacy constraints (i.e., with smaller $\varepsilon$), complying with our theoretical $\tilde{O}(\min(n^2\varepsilon^2, d)$ communication bound.

\paragraph{Organization}
The rest of this paper is organized as follows. We summarize related work in Section~\ref{sec:related} and introduce necessary preliminaries in Section~\ref{sec:preliminary}. We then provide a formal problem formulation in Section~\ref{sec:formulation}. Next, we present and analyze the performance of our main scheme (in terms of privacy, utility, and communication efficiency) and prove its optimality in Section~\ref{sec:worstcase}. After that, we show, in Section~\ref{sec:sparse}, that with additional sparsity assumptions, one can simultaneously reduce the communication cost and increase the accuracy. Finally, we present our experimental results in Section~\ref{sec:experiments} and conclude the paper in Section~\ref{sec:conc}.

\section{Related Work}\label{sec:related}
\paragraph{SecAgg and distributed DP}
SecAgg is cryptographic secure multi-party computation (MPC) that allows the server to collect the sum of $n$ vectors from clients without knowing anyone of them. In our single-server FL setting, SecAgg is achieved via additive masking over a finite group \citep{bonawitz2016practical, bell2020secure}. However, the vanilla FL with SecAgg does not provide provable privacy guarantees since the sum of updates may still leak sensitive information \citep{melis2019exploiting, song2019auditing, carlini2019secret, shokri2017membership}. To address this issue, differential privacy (DP) \citep{dwork2006our}, and in particular, DP-SGD or DP-FedAvg can be employed \cite{song2013stochastic, bassily2014private, geyer2017differentially, mcmahan2017learning}. In this work, we aim to provide privacy guarantees in the form of R\'enyi DP \citep{mironov2017renyi} because it allows for accounting end-to-end privacy loss tightly.

We also distinguish our setup from the local DP setting \citep{kasiviswanathan2011can, evfimievski2004privacy, warner1965randomized}, where the data is perturbed on the client-side before it is collected by the server. Local DP, which allows for a possibly malicious server, is stronger than distributed DP, which assumes an honest-but-curious server. Thus, local DP suffers from worse privacy-utility tarde-offs~\citep{kasiviswanathan2011can,duchi2013local, kairouz16}.

\paragraph{Model compression, sketching, and random projection}
There has been a significant amount of recent work on reducing the communication cost in FL, see \citet{kairouz2019advances}. Among them, popular compression approaches include gradient quantization \citep{alistarh17qsgd, bernstein2018signsgd}, sparsification \citep{aji2017sparse, lin2017deep, wangni2017gradient}, and entropy encoders \citep{havasi2018minimal, oktay2019scalable}. However, since these schemes are mostly non-linear, they cannot be combined with SecAgg where all the encoded messages will be summed together. Therefore, in this work, we resort to compression schemes with \emph{linear} encoders. The only exception are sketching based methods \citep{rothchild2020fetchsgd, haddadpour2020fedsketch}. Our work differs from them in three aspects. First, we consider FL with privacy and SecAgg, whereas \cite{rothchild2020fetchsgd} only aims at reducing communication. Second, although we use the same count-sketch encoder, our decoding method is more aligned with the sparse random projection \citep{kane2014sparser}. In the language of sketching, we decode the sketched model updates by ``count-mean'' instead of count-median, which improves space efficiency, thus requiring less memory to train a real-world large-scale machine learning model. We note that both count-sketch and random projection provides the same worst-case $\ell_2$ error bounds. 

\paragraph{FL with SecAgg and distributed DP}
The closest works to ours are cpSGD~\citep{agarwal2018cpsgd}, DDG~\citep{kairouz2021distributed}, and Skellam~\citep{agarwal2021skellam}, which serve as the main inspiration of this paper. However, all of these methods rely on per parameter quantization and thus lead to $\tilde{\Omega}\lp d \rp$ communication cost. In this work, however, we show that when $d \gg n^2\varepsilon^2$,  we can further reduce dimensionality and achieve the optimal communication cost  $\tilde{O}(n^2\varepsilon^2)$ in this regime. Our scheme also demonstrates $10$x or more compression rates (depending on the privacy budget) relative to the best existing distributed DP schemes. 

\section{Preliminaries}\label{sec:preliminary}
\subsection{Differential Privacy}
We begin by providing a formal definition for $(\varepsilon, \delta)$-differential privacy (DP) \cite{dwork2006calibrating}.
\begin{definition}[Differential Privacy]\label{def:DP}
For $\varepsilon, \delta \geq 0$, a randomized mechanism $M$ satisfies $(\varepsilon, \delta)$-DP if for all neighboring datasets $D, D'$ and all $\mcal{S}$ in the range of $M$, we have that 
$$ \Pr\lp M(D) \in \mcal{S} \rp \leq e^\varepsilon \Pr\lp M(D') \in \mcal{S} \rp+\delta,  $$
where $D$ and $D'$ are neighboring pairs if they can be obtained from each other by adding or removing all the records that belong to a particular user.
\end{definition}
The above DP notion is referred to as user level DP and is stronger than the commonly-used item level DP, where, if a user contributes multiple records, only the addition or removal of one record is protected.

We also make use of Renyi differential privacy (RDP) which allows for tight privacy accounting.
\begin{definition}[Renyi Differential Privacy]\label{def:RDP}
A randomized mechanism $M$ satisfies $(\alpha, \varepsilon)$-RDP if for any two neighboring datasets $D, D'$, we have that $D_\alpha\lp P_{M(D)}, P_{M(D')} \rp\leq \varepsilon$ where $D_{\alpha}\lp P, Q\rp$ is the Renyi divergence between $P$ and $Q$ and is given by 
$$ D_\alpha\lp P, Q \rp \eqDef \frac{1}{\alpha}\log\lp \E_{Q}\lb \lp \frac{P(X)}{Q(X)} \rp^\alpha \rb \rp.$$
\end{definition}
Notice that one can convert $(\alpha, \varepsilon(\alpha))$-RDP to $(\varepsilon_\msf{DP}(\delta), \delta)$-DP using the following lemma from \citet{asoodeh2020better, canonne2020discrete, bun2016concentrated}:
\begin{lemma}\label{lemma:rdp_to_approx_dp}
    If $M$ satisfies $\lp \alpha, \varepsilon \rp$-RDP, then, for any $\delta > 0$, $M$ satisfies $\lp \varepsilon_{\msf{DP}}(\delta), \delta \rp$, where
$$ \varepsilon_{\msf{DP}}(\delta) = \varepsilon+\frac{\log\lp 1/\alpha\delta \rp}{\alpha-1}+\log(1-1/\alpha). $$
\end{lemma}

\subsection{The Distributed Discrete Gaussian Mechanism}\label{sec:ddg}
The previous work of \citet{kairouz2021distributed} proposed a scheme based on the discrete Gaussian mechanism (denoted as $\texttt{DDG}$) which achieves the best mean square error (MSE) $O\lp \frac{c^2d}{n^2\varepsilon^2} \rp$ with a $\frac{1}{2}\varepsilon^2$-concentrated differential privacy guarantee. The encoding scheme mainly consists of the following four steps: (a) scaling, (b) random rotation (therein flattening), (c) conditional randomized rounding, and (d) perturbation, which we summarize in Algorithm~\ref{alg:ddg_simplified} below.

\begin{algorithm}[tbh]
\caption{The DDG mechanism}\label{alg:ddg_simplified}
    \begin{algorithmic}
	\STATE \textbf{Inputs:} Private vector $x_i\in \mbb{R}^d$; clipping threshold $c$;  modulus $M \in \mbb{N}$; noise scale $\sigma > 0$;
	
	\STATE Clip and scale $x_i$ so that $\lV x'_i \rV_2 < c$\;%
	\STATE Randomly rotate vector: $x''_i = U_{\msf{rotate}}\cdot x_i'$%
	\STATE Stochastically round and discretize $x_i''$ into $x_i''' \in \mbb{Z}^d$
	\STATE $Z_i = x'''_i + \mcal{N}_{\mbb{Z}}(0, \sigma^2) \mod{M}$, where $\mcal{N}_{\mbb{Z}}$ is the discrete Gaussian noise\;
	\STATE\textbf{Return:} $Z_i \in \mbb{Z}_M^d$
	\end{algorithmic}
\end{algorithm}

Upon aggregating $\hat{\mu}_{z} = \sum_{i\in[n]} Z_i$, the server can rotate $\hat{\mu}_z$ reversely and re-scale it back to decode the mean $\hat{\mu} =\frac{1}{n}\sum_i x_i$. We refer the reader to Algorithm~\ref{alg:ddg_detailed} in the appendix for a detailed version of DDG. By picking the parameters properly (see Theorem~\ref{thm:ddg}), Algorithm~\ref{alg:ddg_simplified} has the following properties:
\begin{itemize}
\setlength\itemsep{0.1em}
    \item Satisfies $(\alpha, \frac{\varepsilon^2}{2}\alpha)$-RDP, which implies $(\varepsilon_\msf{DP}, \delta)$-DP  with $\varepsilon_\msf{DP} = O_{\delta}\lp \varepsilon^2 \rp$%
    \item Uses  $O\lp d\log\lp n+\sqrt{\frac{n^3\varepsilon^2}{d}}+\frac{\sqrt{d}}{\varepsilon}\rp\rp$ bits of per client
    \item Has an MSE of $ \E\lb \lV \hat{\mu} - \mu \rV^2_2 \rb= O\lp \frac{c^2d}{n^2\varepsilon^2} \rp$.
\end{itemize}

\subsection{Sparse random projection and count-sketch}\label{sec:sparse_random_projection}
We now provide background on sparse random projection~\cite{kane2014sparser} and count-sketching, which allows us to reduce the dimension of local gradients from $\mbb{R}^d$ to $\mbb{R}^m$ with $m \ll d$. These schemes are \emph{linear}, making them compatible with \SA.

Let $S_1,...,S_t \in \lbp -1, 0, 1 \rbp^{d\times w}$ be $t$ identical and independent count-sketch matrices, that is, \begin{equation}\label{eq:S}
    \lp S_i\rp_{j, k} = \sigma_i(j)\cdot\bbm{1}_{\lbp h_i(j)=k\rbp},
\end{equation}
for some independent hash functions $h_i:[d] \ra [w]$ and $\sigma_i:[d]\ra\lbp -1, +1\rbp$. Let $m = t\times w$, then the sparse random projection matrix $S\in \mbb{R}^{d \times m}$ is then defined as stacking $S_1,...,S_t$ vertically, that is,
\begin{equation}\label{eq:def_S}
    S^\intercal = \frac{1}{\sqrt{t}}
\begin{bmatrix}
S^\intercal_1, S^\intercal_2,...,S_t^\intercal
\end{bmatrix}.
\end{equation}

Under this construction, $S$ is sparse in the sense that each column contains exactly $t$ $1$s. We introduce several several nice properties that $S$ possesses which will be used in our analysis.

The following two lemmas controls the distortion of the embedded vector $S\cdot g$ for a $g \in \mbb{R}^d$.
\begin{lemma}\label{lemma:proj_var_bdd}
Let $S$ be defined as in \eqref{eq:def_S}. For any $g_1, g_2 \in \mbb{R}^d$, $\E\lb g_1^\intercal S^\intercal S g_2 \rb = \lan g_1, g_2 \ran$ and
\begin{align*}
    \E_{S}\lb \lan Sg_1, Sg_2\ran^2 \rb \leq \lan g_1, g_2\ran^2 + \frac{2}{m}\lV g_1\rV_2^2\cdot\lV g_2\rV_2^2.
\end{align*}
Furthermore, for any $g \in \mcal{R}^d$,
\begin{align}\label{eq:proj_var_bdd}
    \E_S\lb\lV S^\intercal Sg-g\rV^2_2\rb \leq \frac{2d}{m}\lV g \rV^2_2.
\end{align}
\end{lemma}
The proof follows by directly computing $\E_{S}\lb \lan Sg_1, Sg_2\ran^2 \rb$ (which can be written as a quadratic function of $S$ and $g_1$, $g_2$). See, for instance, Lemma~D.15 in \cite{anonymous2022iterative}.

\begin{lemma}[Sparse Johnson-Lindenstrauss lemma \citep{kane2014sparser}]\label{lemma:SJL}
Let $S$ be defined in \eqref{eq:def_S} and let $g \in \mbb{R}^d$. Then as long as $m \geq \Theta\lp \frac{1}{\alpha^2}\log\lp \frac{1}{\beta} \rp \rp$ and $t \geq \Theta\lp \frac{1}{\alpha} \log\lp \frac{1}{\beta} \rp \rp$, 
\begin{align}
    \Pr\lbp \lV S\cdot g \rV^2_2 \geq (1+\alpha)\lV g\rV^2_2 \rbp \leq \beta.
\end{align}
\end{lemma}

Finally, Lemma~\ref{lemma:inverse_sketch} states that the ``unsketch'' operator preserves the $\ell_2$ norm.
\begin{lemma}\label{lemma:inverse_sketch}
    Let $S, m, t$ be defined in \eqref{eq:def_S} and $v \in \mbb{R}^m$ (which can possibly depends on $S$) with $\E\lb\lV v \rV^2_2| S\rb \leq B^2$ almost surely. Then 
    it holds that $ \E\lb \lV S^\intercal v \rV^2_2\rb \leq \frac{8dB^2}{m}. $
\end{lemma}

\section{Problem Formulation}\label{sec:formulation}

We start by formally presenting the distributed mean estimation (DME) \cite{an2016distributed} problem under differential privacy. Note that DME is closely related to the federated averaging (FedAvg) algorithm \cite{mcmahan2017communication}, where in each round, the server updates the global model using a noisy estimate of the mean of local model updates. Such a noisy estimate is typically obtained via a DME mechanism, and thus one can easily build a DP-FedAvg scheme from a DP-DME scheme.

\paragraph{Distributed mean estimation under differential privacy}
Consider $n$ clients each with a data vector $x_i \in \mathbb{R}^d$ that satisfies $\lV x_i \rV_2 \leq c$ 
(e.g., a clipped local model update). After communicating with $n$ clients, a server releases a noisy estimates $\hat{\mu}$ of the mean $\mu \triangleq \frac{1}{n}\sum_i x_i$, such that 1) $\hat{\mu}$ satisfies a differential privacy constraint (see Definition~\ref{def:DP} and Definition~\ref{def:RDP}), and 2) $\E\lb \lV \hat{\mu} - \mu \rV^2_2 \rb$ is minimized. The goal is to design a communication protocol (which includes local encoders and a central decoder) and an estimator $\hat{\mu}$.

\begin{figure}[ht]
\begin{center}
{\includegraphics[width=0.5\linewidth]{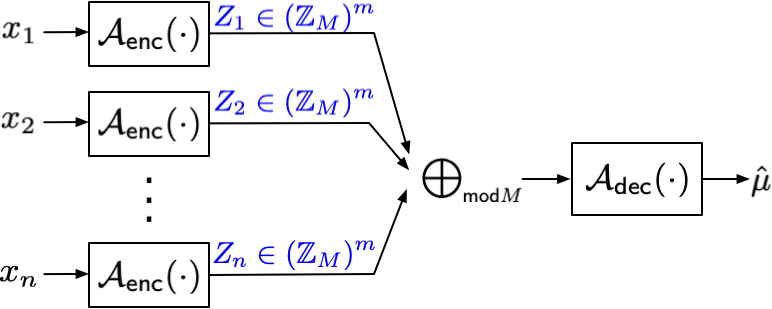}}
\caption{Private mean estimation via SecAgg.}
\label{fig1}
\end{center}
\end{figure}

In this paper, we consider two different DP settings. The first is the \textit{centralized}  DP setting: the server has access to all $x_i$'s, i.e., $\hat{\mu} = \hat{\mu}\lp x_1,...,x_n \rp$.
The  second is the \textit{distributed} DP via \SA setting: under this setting, each client is subject to a $b$-bit communication constraint, so they must first encode $x_i$ into a $b$-bit message, i.e., $Z_i = \mcal{A}_{\msf{enc}}(x_i) \in \mcal{Z}$ with $\lba\mcal{Z}\rba \leq 2^b$ (See Figure~\ref{fig1} for an illustration). However, instead of directly collecting $Z_1,...,Z_n$, the server can only observe the \emph{sum} of them, so the estimator must be a function of $\sum_{i=1}^n Z_i$ (i.e., $\hat{\mu} = \hat{\mu}\lp \sum_{i=1}^n Z_i \rp$). Moreover, we require that the sum $\sum_i Z_i$ satisfies DP (which is stronger than requiring $\hat{\mu}$ to be DP), meaning that individual information will not be disclosed to the server as well. Notice that since \SA operates on a finite additive group, we require $\mcal{Z}$ to have an additive structure. Without loss of generality, we will set $\mcal{Z}$ to be $\lp\mbb{Z}_M\rp^m$ for some $m, M \in \mbb{N}$, where $\mbb{Z}_M$ denotes the group of integers modulo $M$ (equipped with modulo $M$ addition) and $m$ is the dimension of the space we are projecting onto. In other words, we allocate $\log M$ bits for every coordinate of the projected vector. Note that in this case, the total per-client communication cost is $b = m\log M$.%

For a fixed privacy constraint, the fundamental problem we seek to solve is: what is the smallest communication cost under the distributed DP setting needed to achieve the accuracy of the centralized DP setting? Further, we seek to discover schemes that (a) achieve the optimal privacy-accuracy-communication trade-offs and (b) are memory efficient and computationally fast in encoding and decoding.

\section{Communication Cost of DME with SecAgg}

In this section, we characterize the \emph{optimal communication cost} under the distributed DP via SecAgg setting, defined as the smallest number of bits (as a function of $n, d, \varepsilon$) needed to achieve the same accuracy (up to a constant factor) of the centralized setting under the same $(\varepsilon, \delta)$-DP constraint.

\paragraph{Optimal MSE under central DP} 

We start by specifying the optimal accuracy under a fully trusted server and no communication constraints. Under a DME setting where $\lV x_i \rV_2 \leq c$ for all $i\in[n]$, the $\ell_2$ sensitivity of the mean query $\mu(x_1,...,x_n) \eqDef \frac{1}{n}\sum_{i=1}^n x_i$ is bounded by 
$$ \Delta\lp \mu \rp \eqDef \max_{x^n, x_1'} \lV \mu(x_1,...,x_n) - \mu(x_1',...,x_n) \rV_2^2 \leq \frac{2c^2}{n}.  $$
Therefore, to achieve ($\varepsilon, \delta)$ central DP, the server can add coordinate-wise independent Gaussian noise to $\mu$. This gives an $\ell_2$ error that scales as 
$O_{\delta}\lp \frac{c^2d}{n^2\varepsilon^2} \rp$, which is known to be tight in the high privacy regime \cite{kamath2020primer}. Moreover, the resulting estimator is \emph{unbiased}\footnote{Notice that in this work, we are mostly interest in unbiased estimators (see Remark~\ref{rmk:unbiasedness} for a discussion).}.

\paragraph{Communication costs of DDG}
Next, we examine the communication costs of previous distributed DP schemes such as the DDG mechanism. As mentioned in Section~\ref{sec:ddg} and Theorem~\ref{thm:ddg}, in order to achieve the $O_{\delta}\lp \frac{c^2d}{n^2\varepsilon^2} \rp$ error, the communication cost of DDG must to be at least $\Theta\lp d \log\lp d/\varepsilon^2 \rp \rp$ bits. Note that the communication cost scales up with $1/\varepsilon$ because in the high-privacy regimes, the noise variance needs to be increased accordingly to provide stronger DP. Thus SecAgg's group size needs to be enlarged to capture the larger signal range and avoid catastrophic modular clipping errors. A similar phenomenon occurs for other additive noise-based mechanisms, e.g, the Skellam and binomial \cite{agarwal2018cpsgd, agarwal2021skellam} mechanisms.

However, we show that the $\Theta\lp d \log\lp d/\varepsilon^2 \rp \rp$ cost is strictly sub-optimal. In particular, the linear dependency on $d$ can be further improved when $n^2\varepsilon^2 \ll d$. To demonstrate this, we start by the following example to show that under a $(\varepsilon, \delta)$ central DP constraint, one can reduce the dimensionality to $m = O(n^2\varepsilon^2)$ without harming the MSE.

\paragraph{Dimensionality reduction under central DP} Consider the following simple project-and-perturb mechanism:

\begin{enumerate}
    \item The server generates a random projection matrix $S\in \mbb{R}^{m\times d}$ according to the sparse random projection defined in Section~\ref{sec:sparse_random_projection} and broadcasts it to $n$ clients.
    \item Each client sends $ y_i \eqDef \msf{clip}_{\ell_2, 1.1c}\lp S x_i \rp$.
    \item The server computes $\hat{\mu} \eqDef S^\intercal\lp \frac{1}{n}\sum_i y_i+N\lp 0, \sigma^2\mbb{I}_m \rp\rp$,  where $\sigma^2 = \Theta\lp \frac{c^2}{n^2\varepsilon^2}\rp$.
\end{enumerate}

We claim that the above project-and-perturb approach satisfies $(\varepsilon, \delta)$ DP and achieves the optimal MSE order. In other words, we can reduce the dimensionality \emph{for free}. 

To see why this is true, observe that we can decompose the overall $\ell_2$ error $\lV \hat{\mu} - \mu\rV^2_2$ into three parts: (1) the clipping error (i.e., $\lV y_i - Sx_i \rV^2_2$), (2) the compression error (i.e. $\lV\mu - S^\intercal S\lp\sum_i x_i/n\rp\rV^2_2$), and (3) the privatization error $\lV S^\intercal N(0, \sigma^2 \mbb{I}_m)\rV^2_2$. Then, we argue that all of them have orders less than or equal to $O\lp \frac{c^2d}{n^2\varepsilon^2} \rp$, as long as we select $m = \Theta(n^2\varepsilon^2)$ and $t = \Theta \lp \log d+ \log(n^2\varepsilon^2) \rp$.

First, the clipping error is small since the random projection $S$ satisfies the Johnson-Lindenstrauss (JL) property (see Lemma~\ref{lemma:SJL} in the appendix), which implies that $\lV S x_i\rV_2 \approx \lV x_i\rV_2 \leq c$ and that the clipping happens with exponentially small probability. Second, Lemma~\ref{lemma:proj_var_bdd} (in the appendix) suggests that compression error scales as $O\lp \frac{c^2d}{m} \rp$. Thus by picking $m = \Theta\lp n^2\varepsilon^2 \rp$, we ensure the compression error to be at most $O\lp \frac{c^2d}{n^2\varepsilon^2} \rp$. Finally, since the Gaussian noise $N$ added at Step 3 is independent of $S$, the privatization error can also be bounded by $O\lp \frac{c^2d}{n^2\varepsilon^2} \rp$.

We summarize this in the following lemma, where the formal proof is deferred to Section~\ref{proof:utility_cdp} in the appendix.

\begin{lemma}\label{lemma:utility_cdp}
Assume $\lV x_i \rV_2 \leq c$ for all $i \in [n]$. Then the output of the above mechanism $\hat{\mu}$ satisfies $\lp \varepsilon, \delta \rp$-DP. Moreover, if
$S$ (defined in \eqref{eq:def_S}) is generated with $m = \Theta(n^2\varepsilon^2)$ and $t = \Theta \lp \log d+ \log(n^2\varepsilon^2) \rp$, it holds that $\E\lb \lV \hat{\mu} - \mu \rV^2_2  \rb = O\lp\frac{c^2d }{n^2\varepsilon^2}\rp$.

\end{lemma}

\paragraph{Dimensionality reduction with SecAgg}
The above example shows that with a random projection, we can reduce the dimensionality from $d$ to $O(n^2\varepsilon^2)$ without increasing the MSE too much. Thus, under the distributed DP via SecAgg setting, we combine random projection with the DDG mechanism to arrive at our main scheme in Algorithm~\ref{alg:sketch_mean_decoder}. 

\begin{algorithm}[tb]
\caption{Private DME with random projection}\label{alg:sketch_mean_decoder}
\begin{algorithmic}
   \STATE {\bfseries Input:} Cleints' data $  x_1,...,x_n \in \mb{B}_{d}(c)$, compression parameter $m\in\mbb{N}$
   \STATE The server generates a sketching matrix $S \in \mbb{R}^{m\times d}$ \;
\STATE The server broadcasts $S$ to all clients\;
    
    \FOR{$i \in [n]$}
	\STATE Client $i$ computes $y_i \eqDef S x_i$ and $Z_i \eqDef \texttt{DDG}_{\msf{enc}}\lp y_i \rp \in \mbb{Z}_M^m$ with $\ell_2$ clipping parameter $1.1c$ (and other parameters being the same as in Algorithm~\ref{alg:ddg_simplified} with $d$ being replaced by $m$)\;
	\ENDFOR
    \STATE The server aggregates $Z_1,...,Z_n$ with SecAgg and decodes $\hat{\mu}_y = \frac{1}{n}\texttt{DDG}_{\msf{dec}}\lp\sum_{i\in[n]} Z_i\rp$\;
    \STATE The server computes $ \hat{\mu} = S^\intercal \hat{\mu}_y$.
    \STATE {\bfseries Return:} $\hat{\mu}$
\end{algorithmic}
\end{algorithm}

To control the $\ell_2$ error of Algorithm~\ref{alg:sketch_mean_decoder}, we adopt the same strategy as in Lemma~\ref{lemma:utility_cdp} (i.e., decompose the end-to-end error into three parts), with the privatization error being replaced by the error due to DDG. However, this will cause an additional challenge, as the error due to DDG is no longer independent of $S$ (as opposed to the Gaussian noise in the previous case). To overcome this difficulty, we leverage the fact that \emph{for any} projection matrix $S$, the (expected) $\ell_2$ error is bounded by $O\lp \frac{c^2m}{n^2\varepsilon^2} \rp$ and develop an upper bound on the final MSE accordingly (see Section~\ref{proof:random_projection_upper_bound} for a formal proof).

We summarize the performance guarantees of Algorithm~\ref{alg:sketch_mean_decoder} in the following theorem.
\begin{theorem}\label{thm:random_projection_upper_bound}
Let $S$ in Algorithm~\ref{alg:sketch_mean_decoder} be generated according to \eqref{eq:def_S} with $m = \Theta(n^2\varepsilon^2)$ and $t = \Theta \lp \log d+ \log(n^2\varepsilon^2) \rp$. Assume $\lV x_i \rV \leq c $ for all $i \in [n]$. Then as long as $n^2\varepsilon^2 \leq d$, the following holds:
\begin{itemize}
    \item Algorithm~\ref{alg:sketch_mean_decoder} satisfies $\lp \alpha, \frac{\varepsilon^2}{2}\alpha\rp$-RDP,
    \item the MSE is bounded by
    $ \E\lb \lV \hat{\mu} - \mu \rV^2_2 \rb = O\lp \frac{c^2d}{n^2\varepsilon^2} \rp, $
    \item the per-client communication is {\small
    $$ O\lp m\log\lp n+\sqrt{\frac{n^3\varepsilon^2}{m}}+\frac{\sqrt{m}}{\varepsilon} \rp\rp = O\lp n^2\varepsilon^2\log n \rp. $$}
\end{itemize}
\end{theorem}

\paragraph{Lower bounds for private DME with SecAgg}
Next, we complement our achievability result in Theorem~\ref{thm:random_projection_upper_bound} with a matching communication lower bound. Our lower bound indicates that Algorithm~\ref{alg:sketch_mean_decoder} is indeed optimal (in terms of communication efficiency), hence characterizing the fundamental privacy-communication-accuracy trade-offs. 

The lower bound leverages the fact that under \SA, the \emph{individual} communication budget each client has is equal to the \emph{total} number of bits the server can observe, which are all equal to the cardinality of the finite group $\mcal{Z}$ that \SA acts on. Therefore, even if each client sends a $b$-bit message (so the total information transmitted is $n\cdot b$ bits), the server still can only observe $b$ bits information. 

With this in mind, to give a per-client communication lower bound under \SA, it suffices to lower bound the total number of bits needed for reconstructing a $d$-dim vector (i.e. the mean vector $\mu$) within a given error (i.e. $O_\delta\lp c^2d/(n^2\varepsilon^2) \rp$). Towards this end, we first derive a general lower bound that characterizes the communication-accuracy trade-offs for compressing a \emph{single} $d$-dim vector in a centralized setting.

\begin{theorem}[compression lower bounds]\label{thm:compression_lb}
Let $v \in \mbb{R}^d$ and $\lV v \rV_2 \leq c$. Then for any (possibly randomized) compression operator $\mcal{C}: \mb{B}_{d}(c) \ra [2^b]$ that compresses $v$ into $b$ bits and any (possibly randomized) estimator $\hat{v}: [2^b] \ra \mb{B}_{d}(c)$, it holds that
\begin{equation}\label{eq:compression_lb_general}
    \min_{\lp \mcal{C}, \hat{v} \rp} \max_{v \in \mb{B}_d(c)} \E\lb \lV \hat{v}\lp \mcal{C}(v) \rp - v \rV^2_2 \rb \geq  c^22^{-2b/d}.
\end{equation}
Moreover, for any unbiased compression scheme (i.e., $\lp \mcal{C}, \hat{v} \rp$ satisfying $ \E\lb \hat{v}\lp \mcal{C}(v) \rp \rb = v$ for all $v \in \mb{B}_d(c)$) with $b < d$,  it holds that
\begin{equation}\label{eq:compression_lb_unbiased}
    \min_{\lp \mcal{C}, \hat{v} \rp} \max_{v \in \mb{B}_d(c)} \E\lb \lV \hat{v}\lp \mcal{C}(v) \rp - v \rV^2_2 \rb \geq  C_0c^2d/b,
\end{equation}
where $C_0 > 0$ is a universal constant.
\end{theorem}

Theorem~\ref{thm:compression_lb}, together with the fact that the per-client bit budget is also the total amount of information the server can observe, we arrive at the following lower bound.

\begin{corollary}
Consider the private DME task with SecAgg as described in Figure~\ref{fig1}. For any encoding function $\mcal{A}_{\msf{enc}}(\cdot)$ with output space $ \mcal{Z} $, if the $\ell_2$ estimation error $\E\lb \lV \hat{\mu} - \mu \rV^2_2 \rb \leq \xi$ for all possible $x_1,...,x_n \in \mb{B}_d(c)$, %
then it  must hold that
$ \log\lba  \mcal{Z}  \rba = \Omega\lp d\log\lp c^2/\xi \rp\rp. $
In addition, if $\hat{\mu}$ is unbiased and $\xi \leq c^2$, then
$ \log\lba  \mcal{Z}  \rba = \Omega\lp dc^2/\xi\rp. $
\end{corollary}

Finally, by plugging $\xi = O\lp \frac{c^2d}{n^2\varepsilon^2} \rp$ (which is the optimal $\ell_2$ error for the mean estimation task under centralized DP model), we conclude that 
\begin{itemize}
    \item $\Omega\lp \max\lp d\log\lp {n^2\varepsilon^2}/{d} \rp, 1\rp \rp$ bits of communication are necessary for general (possibly biased) schemes
    
    \item $\Omega\lp \min\lp n^2\varepsilon^2, d\rp \rp$ bits of communication are necessary for unbiased schemes.
\end{itemize}

We remark that the above lower bounds are both tight but in different regimes. Specifically, the first lower bound, which measures the accuracy in MSE, is tight for small $d$ but is meaningless in high-dimensional or high-privacy regimes where $d\gg n^2\varepsilon^2$. This also implies that $\tilde{\Omega}(d)$ bits are necessary for $d\ll n^2\varepsilon^2$ and that there is no room for improvement on DDG in this regime. On the other hand, the second bound is useful when $d = \Omega\lp n^2\varepsilon^2 \rp$ (with an additional unbiasedness assumption). This is a more practical regime for FL with SecAgg, and our scheme outperforms DDG in this scenario. %

\begin{remark}\label{rmk:unbiasedness}
Notice that in this work, we are mostly interested in unbiased estimators due to the following two reasons: 1) it largely facilities the convergence analysis of the SGD based methods, as these types of stochastic first-order methods usually assume access to an unbiased gradient estimator in each round. 2) In the high-dimensional or high-privacy regimes where $d\gg n^2\varepsilon^2$, the MSE is not the right performance measure since an estimator can have a large bias while still achieving a relative small MSE. 
For instance, in the regime where $n^2\varepsilon^2 \leq d$, the Gaussian mechanism has MSE $\Theta\lp \frac{c^2d}{n^2\varepsilon^2} \rp$; on the other hand, the trivial estimator $\hat{\mu} = 0$ achieves a smaller MSE, equal to $c^2 \leq O\lp\frac{c^2d}{n^2\varepsilon^2}\rp$, but such estimator gives no meaningful information. In order to rule out these impractical schemes, we hence impose the unbiasedness constraint.
\end{remark}

\label{sec:worstcase}

\section{Sparse DME with \SA and DP}
Theorem~\ref{thm:random_projection_upper_bound} in Section~\ref{sec:worstcase} specifies the optimal trade-offs of private DME \emph{for all} possible datasets $x^n$. In other words, it provides a \emph{worst-case} (over all possible $x_i$) bound on the utility and shows that Algorithm~\ref{alg:sketch_mean_decoder} is worst-case optimal. However, we show in this section that with additional assumptions on the data, it is possible to improve the trade-offs beyond what is given in Theorem~\ref{thm:random_projection_upper_bound}. 

One such assumption is \emph{sparsity} of the data, which is justified by several empirical results that gradients tend to be (or are close to being) sparse. We hence study the sparse DME problem, which is formulated as in Section~\ref{sec:formulation} but with an additional $s$-sparsity assumption on $\mu$, i.e., $\lV \mu\rV_0 \leq s$. We present a sparse DME algorithm adapted from Algorithm~\ref{alg:sketch_mean_decoder}, showing that by leveraging the sparse structure of data, one can surpass the lower bound in Theorem~\ref{thm:random_projection_upper_bound}. Moreover, the dependency of communication cost and MSE on the model size $d$ becomes logarithmic.

\paragraph{DME via compressed sensing}
We adopt the same strategy as in Section~\ref{sec:worstcase}, (i.e., use a linear compression scheme to reduce dimensionality). However, instead of applying the linear decoder $S^\intercal\hat{\mu}_y$, we perform a more complicated compressed sensing decoding procedure and solve a (regularized) linear inverse problem. Specifically, we modify Algorithm~\ref{alg:sketch_mean_decoder} in the following way:
\begin{enumerate}
    \item For local compression, we replace the sparse random projection matrix $S\in \mbb{R}^{m\times d}$ with an $s$-RIP\footnote{See Definition~\ref{def:soft_re} for a weaker definition.} matrix (in particular, we use a Gaussian ensemble, i.e., $S_{i,j} \diid N(0, 1)$) and set $m = O\lp s\log d \rp$.
    \item To decode $\hat{\mu}$,  the server solves the following $\ell_1$ regularized linear inverse problem \eqref{eq:lasso} (i.e., LASSO \citep{tibshirani1996regression})
    where the $\lambda_n$ is set to be of the order $O\lp \frac{c\log d}{n\varepsilon}\rp$.
\end{enumerate}
With the above modifications, we arrive at Algorithm~\ref{alg:sparse_DME_lasso}:
\begin{algorithm}[ht]
    \begin{algorithmic}
	\STATE \textbf{Inputs:} clients' data $x_1,...,x_n$, sparse parameter $s$
	
	\STATE The server generates an compression matrix $S \in \mbb{R}^{m\times d}$ with $m = \Theta\lp s\log d \rp$ and $S_{i,j}\diid N(0, 1) $ and computes its largest singular value $\sigma_\msf{max}(S)$\;
	\STATE The server broadcasts $S, \sigma_\msf{max}(S)$ to all clients\;
	\FOR{$i \in [n]$}
	\STATE Client $i$ computes $y_i \eqDef S x_i$ and $z_i \eqDef \texttt{DDG}_{\msf{enc}}\lp y_i \rp \in \mbb{Z}_M^m$ with clipping rate $c' = c\sigma_\msf{max}(S)$ and dimension $d' = m$\;
	\ENDFOR
	
    \STATE The server aggregates $Z_1,...,Z_n$ with SecAgg and decodes $\hat{\mu}_y = \frac{1}{n}\texttt{DDG}_{\msf{dec}}\lp\sum_{i\in[n]} Z_i\rp$\;
    \STATE The server solves the following Lasso:
    \begin{equation}\label{eq:lasso}
        \hat{\mu} \in \arg\min_{x \in \mbb{R}^d}\lbp \frac{1}{m}\lV \hat{\mu}_y-Sx \rV_2+\lambda_n\lV x \rV_1 \rbp, 
    \end{equation} 
    where the regularization is picked to satisfy 
    $\lambda_n = O\lp \frac{c\log d}{n\varepsilon}\rp.$
	\STATE \textbf{Return: }{$\hat{\mu}$}
	\end{algorithmic}
	\caption{sparse DME via Compressed Sensing}\label{alg:sparse_DME_lasso}
\end{algorithm}

The next theorem provides the privacy and utility guarantees of Algorithm~\ref{alg:sparse_DME_lasso}.
\begin{theorem}[Sparse private DME]\label{thm:compressed_sensing_ddp}
Algorithm~\ref{alg:sparse_DME_lasso} satisfies $(\alpha, \frac{1}{2}\varepsilon^2\alpha)$-RDP. In addition, if $\lV \mu\rV_0 \leq s$, it holds that
\begin{itemize}
    \item the per-client communication cost is $O\lp s\log d \log \lp n^2 +{s\log d}/{\varepsilon^2}\rp\rp$;
    \item the MSE is bounded by $ O\lp \frac{c^2s\log^2 d}{n^2\varepsilon^2} \rp$.
\end{itemize}
\end{theorem}

Observe that under sparsity, both the accuracy and the communication cost depend on $d$ logarithmically. This implies that by leveraging the sparsity, we can replace $d$ with an ``effective'' dimension of $s\log d$. However, Algorithm~\ref{alg:sparse_DME_lasso} is more complicated than Algorithm~\ref{alg:sketch_mean_decoder} as it requires tuning hyper-parameters such as $s$ and $\lambda_n$.

\begin{remark}
The communication cost in Theorem~\ref{thm:compressed_sensing_ddp} no longer depends only on $n\varepsilon$, thus exhibiting a different behavior from the non-sparse case (i.e., that of  Theorem~\ref{thm:random_projection_upper_bound}). We remark that this is because we only present the result for the $s\log d \ll n^2\varepsilon^2$ (which is more reasonable in practice). However, one can extend the similar analysis in Section~\ref{sec:worstcase} and obtain the results for $s\log d \gg n^2\varepsilon^2$ regime.
\end{remark}

\label{sec:sparse}

\section{Empirical Analysis}\label{sec:experiments}
We run experiments on the full Federated EMNIST and Stack Overflow datasets~\citep{caldas2018leaf}, two common benchmarks for \FL tasks. F-EMNIST has $62$ classes and $N=3400$ clients with a total of $671,585$ training samples. Inputs are single-channel $(28,28)$ images. The Stack Overflow (SO) dataset is a large-scale text dataset based on responses to questions asked on the site Stack Overflow. The are over $10^8$ data samples unevenly distributed across $N=342477$ clients. We focus on the next word prediction (NWP) task: given a sequence of words, predict the next words in the sequence. On both datasets, we select $n\in[100,1000]$ and $R=1500$. On F-EMNIST, we experiment with a $\approx 10^6$ parameter (4 layer) Convolutional Neural Network  (CNN) used by~\citet{kairouz2021distributed}. On SONWP, we experiment with a $\approx 4\cdot10^6$ parameter (4 layer) long-short term memory (LSTM) model---the same as prior work~\cite{andrew2019differentially,kairouz2021distributed}.
In both cases, clients train for $1$ local epoch using SGD. Only the server uses momentum.
For \DDP, we use the geometric adaptive clipping of~\cite{andrew2019differentially}.
We use the same procedure as~\citet{kairouz2021distributed}. We randomly rotate vectors using the Discrete Fourier Transform. We use their hyperparameter values for conditional randomized rounding and modular clipping. We communicate $16$ bits per parameter for F-EMNIST and $18$ for SONWP unless otherwise indicated.
We repeat all experiments with $5$ different seeds (or more, where stated).
We provide full detail on the models, datasets, and training setups in Appendix~\ref{app:models}, as well as the chosen values of the noise multiplier and sparse random projections (via sketching) parameters in Appendix~\ref{app:DP}. We provide details of our algorithms for sparse random projections (via sketching) in Appendix~\ref{app:sketching-practice}.

\begin{figure}[t!]
    \centering
    \captionsetup[subfigure]{width=0.95\linewidth}
    \begin{subfigure}[t]{0.48\linewidth}
        \includegraphics[width=0.97\linewidth]{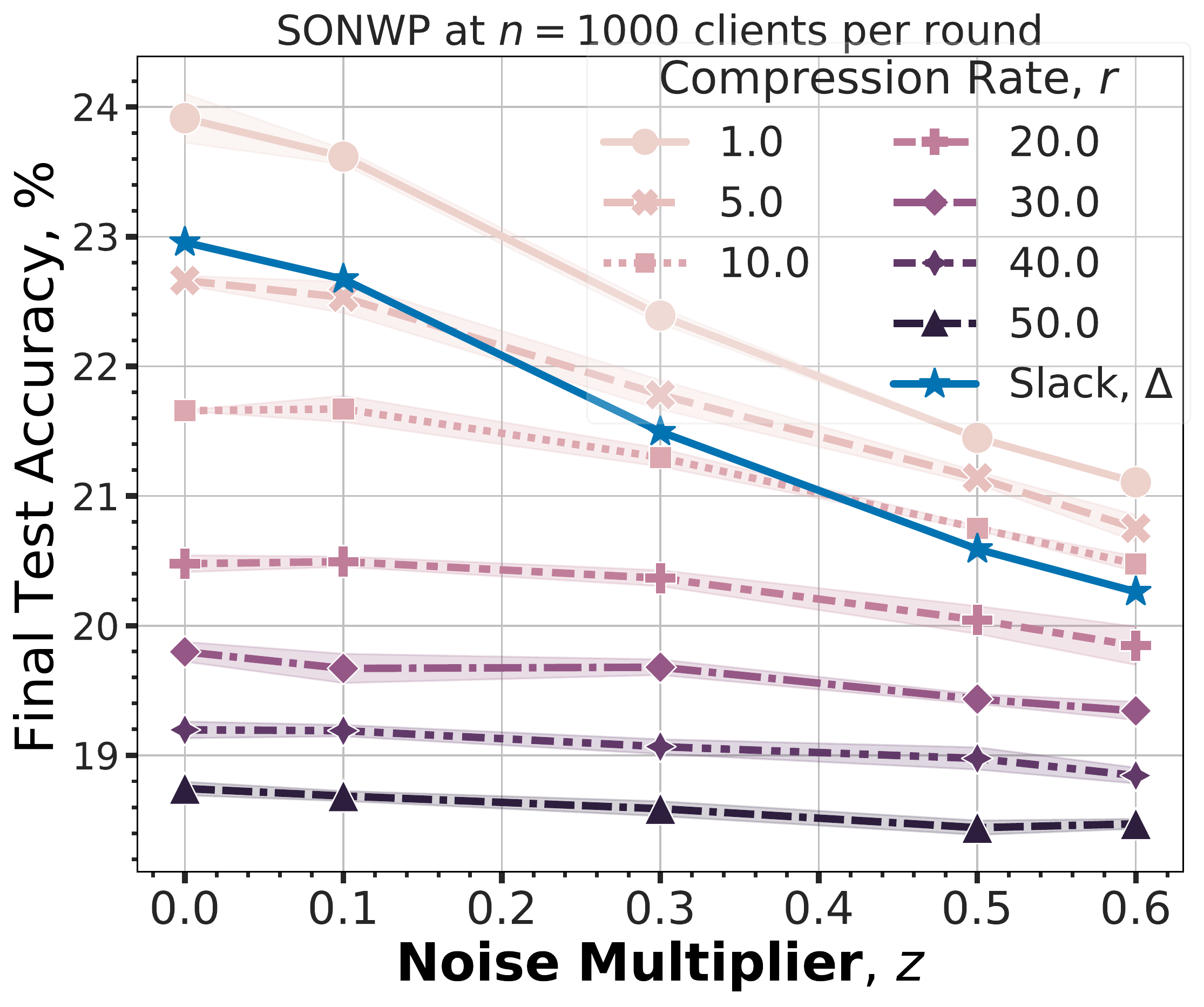}
        \caption{\textbf{At $\mathbf{z\geq0.5}$, we attain $r\geq\mathbf{10}$x on SONWP} and without DP ($z=0$), $\approx4$x. $\Delta=4\%$. }
        \label{fig:sonwp}
    \end{subfigure}
    \begin{subfigure}[t]{0.48\linewidth}
        \includegraphics[width=\linewidth]{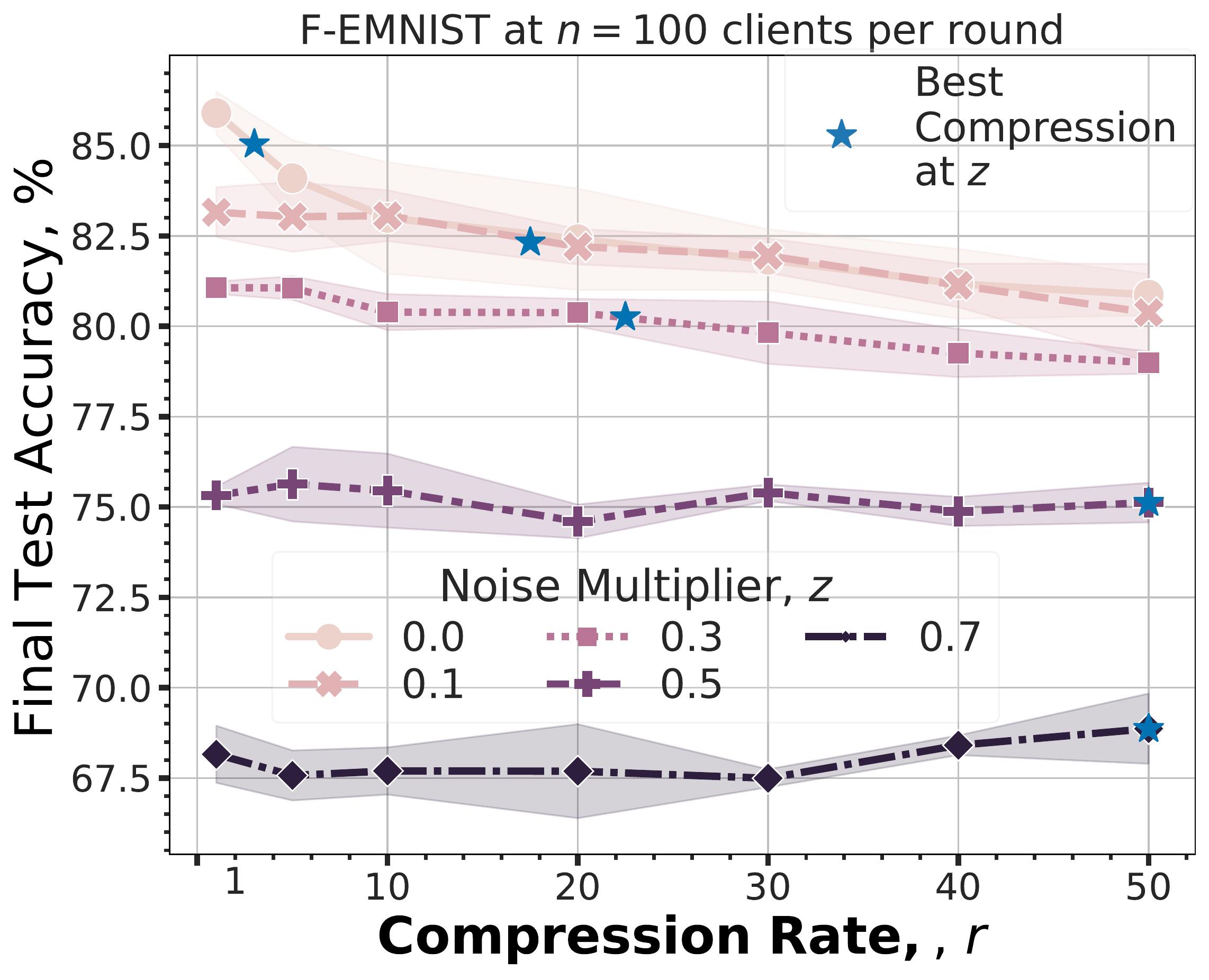}
        \caption{\textbf{At $\mathbf{z\geq0.5}$ we attain $r\geq\mathbf{50}$x on F-EMNSIT}. Without DP, we cannot significantly compress. $\Delta=1\%$.}
        \label{fig:femnist}
    \end{subfigure}
    \caption{\textbf{Higher privacy requirements lead to higher attained compression rates with a slack of $\Delta$.} A higher (fixed) compression rate can also attain tighter privacy at no cost in performance.}
    
    \label{fig:puc}
\end{figure}

\paragraph{Analyzing the \puc tradeoff}\label{ssec:more-privacy-compression}
We now study the best compression rates that we can attain without significantly impacting the current performance.
For this, we train models that achieve state-of-the-art performance on the FL tasks we consider, and allow a slack of $\Delta=4\%$ relative to these models when trained without compression ($r=1$x). We first consider $z=0$, i.e., no DP, and see that about $r=4$x compression can be attained. However, recall that our theoretical results suggest that as $z$ increases and thus $\varepsilon$ decreases, fewer bits of communication are needed. Our experiments echo this finding: at $z=0.3$, we observe $r=5$x is now attainable and at $z\geq0.5$, $r=10$x is attainable. The highest $z$ we display can correspond both to a tighter privacy regime ($\varepsilon \approx 10$ or less) but, importantly also to models that are still highly performant indicating that a practitioner could reasonably select both these $z$ and $r$ to train models comparable to the state-of-the-art.
Our experiments on F-EMNIST in Figure~\ref{fig:femnist} show similar findings where fixing $r\geq50$x can attain $z\geq0.5$ for `free'. 
Finally, these results also corroborate that our bound of $\tilde{O}\lp \min(n^2\varepsilon^2, d) \rp$ is significantly less ($\approx 10$x) than that of~\citet{kairouz2021distributed} in practice.

Finding that we can significantly compress our models, another question that can be asked is `could a smaller model have been used instead?' To investigate this, we train a smaller model (denoted `small') which has only $\approx 2 \cdot 10^5$ parameters, which is comparable in size to the original CNN model updates compressed by $r\approx 5$x. We observe that this model has significantly lower performance ($>5pp$) across all privacy budgets when we compare them for a fixed latent dimension of $length \times width$. When we compress our original model beyond $r=5$x (smaller than the `small' model), we find that it still significantly outperforms it. These results indicate that training larger models do in fact attain higher performance even for the same latent update size: further, our results indicate we can enable training these larger models under the same fixed total communication.

\paragraph{Quantization or dimensionality reduction?}
Though we achieve significant compression rates from our linear dimensionality reduction technique, we now explore how these results compare with compression via quantization. Theoretically our bound can achieve a minimum compression independent of the ambient gradient size $d$. Because of this, we also expect our methods to outperform those based on quantization because their communication scales with $d$.

\begin{figure}[h]
    \centering
    \includegraphics[width=0.6\linewidth]{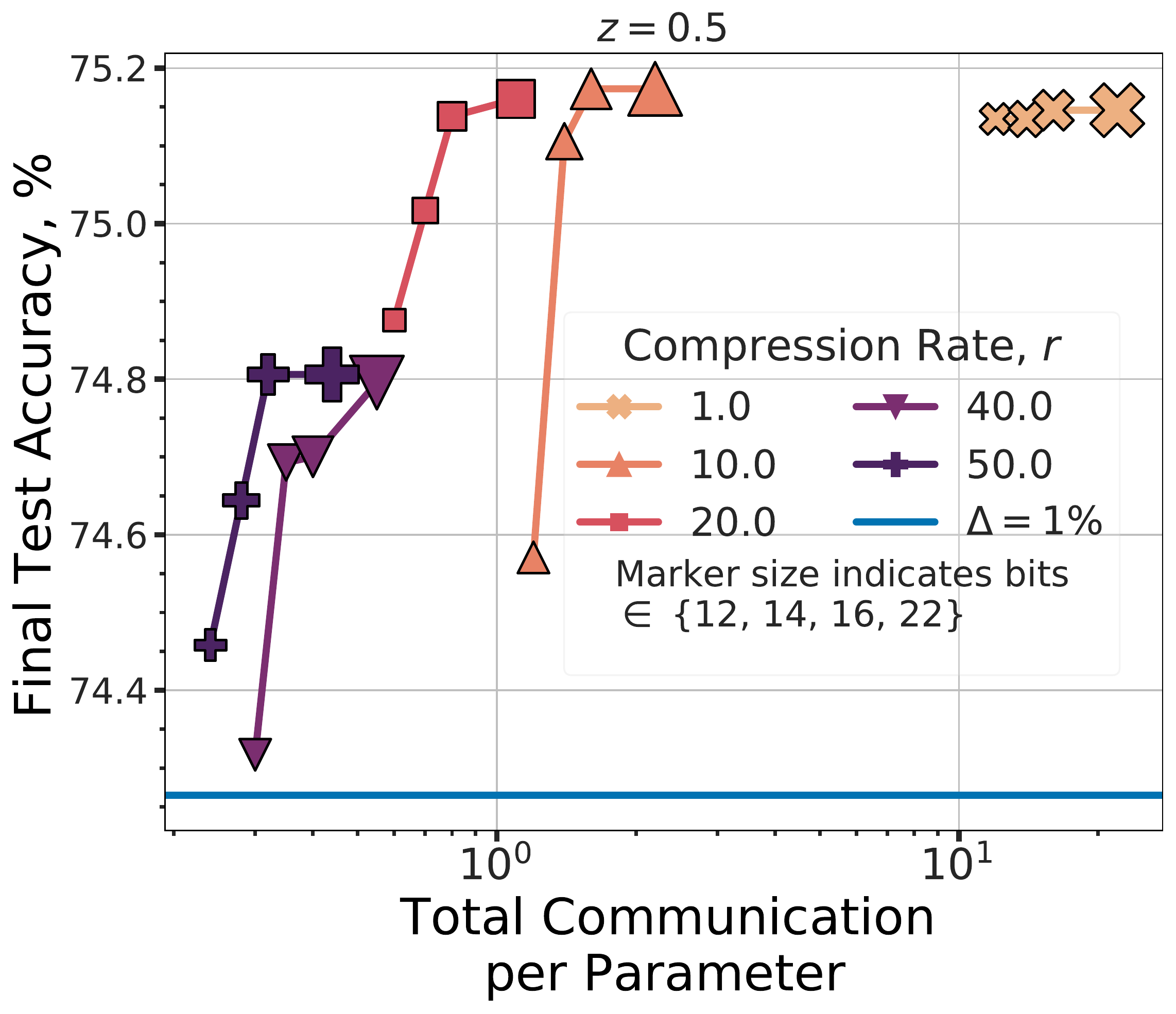}

    \caption{\textbf{Optimizing both $r$ and $b$ can further decrease communication}, to $0.24$ bits per parameter at $z=0.5$. See Figure~\ref{fig:femnist-short} of Appendix~\ref{app:figures} for $z=0.1,0.3$. Note that sometimes at higher bitwidths we observe lower performance within statistical error (standard deviation $\approx 1$)---here, we threshold to the highest accuracy of lower bitwidths to ease visualization. Full results without thresholding and all $r$ are in Figure~\ref{fig:femnist-long} of Appendix~\ref{app:figures}. These results used $30$ different seeds.}
    \label{fig:quant-vary}
\end{figure}

Our results in Figure~\ref{fig:quant-vary} and Table~\ref{tab:sonwp-b-vary} corroborate this hypothesis. We compare our compression against conditional randomized rounding to an integer grid of field size $2^b$. We vary the quantization (bits) per parameter $b$ while allowing the same $\Delta$ as above. Combined with dimensionality reduction, this gives a total communication per ambient parameter as $b/r$. When we use the vanilla distributed DP scheme of~\citet{kairouz2021distributed}, we find that we can only compress to $10$ bits per parameter on SONWP; if we instead favour our linear dimensionality reduction technique, we achieve a much lower $1.2$ bits per parameter (with $r=10$x and $b=12$). For F-EMNIST, we find we can compress down to $0.24$ bits per parameter at $z=0.5$ by optimizing both $b$ and $r$; this is much lower than using only quantization ($10$ bits per parameter) and a marginal increase over only dimensionality reduction ($0.27$). We observe $0.7$ bits per parameter at $z=0.3$ and $1.2$ at $z=0.1$. See Figure~\ref{fig:femnist-short} for results with $z=0.1,0.3$ and Figure~\ref{fig:femnist-long} for the comprehensive results at many $r$ and $b$, both in Appendix~\ref{app:figures}.
\begin{table}[h]
    \centering
    \begin{tabular}{c || c | c | c | c}
        \makecell{\small Noise\\ \small Multiplier, $z$ }& \makecell{\small Chosen Compression\\\small Rate $r$} & \makecell{\small Lowest Bit Width\\\small$b$ per Parameter} & \makecell{\small Total Communication\\\small Per Parameter} & \makecell{\small Final Test\\\small Performance, \%} \\
        \hline\hline
        \multirow{2}{*}{0.3} &
        1 & 10 & 10 & $21.90\pm 0.09$\\
        & \textbf{5} & \textbf{12} & \textbf{2.4} & $\mathbf{21.68\pm0.02}$\\
        \hline
        \multirow{3}{*}{0.5} &
        1 & 10 & 10 & $21.21 \pm 0.18$ \\
        & 5 & 12 & 2.4 & $21.11 \pm 0.08$ \\
        & \textbf{10} & \textbf{12} & \textbf{1.2} & $\mathbf{20.74 \pm 0.05}$ \\
        \hline
    \end{tabular}
    \caption{\textbf{Optimal compression can be found by tuning both the bit width $b$ and compression rate $r$.} Results for SONWP with 1000 clients. We find that increasing $r$ instead of $b$ achieves the highest total compression in all cases. Bold rows show the optimal compression parameters for the given $z$. We note that we cannot set lower than $b=10$ for this setting because SecAgg requires at least $O(\log{(n)}$ bits. The results in the final column take the form mean$\pm$standard deviation.}
    \label{tab:sonwp-b-vary}
\end{table}

Because of this, we now explore if raising the $b$ (decreasing quantization), past the max $b=18$ bits per parameter that we have considered thus far, will decrease the total communication. Inspecting $b=18\to22$ in Figure~\ref{fig:quant-vary} we do observe a marginal increase in test performance with increasing $b$ in some cases, indicating there may be potential to increase $r$. However, we do not find that we can significantly increase $r$ in these cases (see Figure~\ref{fig:femnist-long} of Appendix~\ref{app:figures}).

\paragraph{Impact of cohort size} 

We conduct experiments under (approximately, because varying $n$ impacts $\varepsilon$) fixed $\varepsilon$ to investigate how the cohort size $n$ impacts compression. From Theorem~\ref{thm:compression_lb}, we expect to see that as $n$ increases, so does does communication. Our empirical results closely match, shown in Figure~\ref{fig:n-sonwp} of Appendix~\ref{app:figures}. Further, we observe from Table~\ref{tab:n-emnist} of Appendix~\ref{app:figures} that under sufficiently high $z$, we can increase $n$ while keeping the total message size fixed (increase $r$ by $d/(\frac{n_2}{n_1}\frac{\log{n_2}}{\log{n_1}})$, where $n_2>n_1$) to \emph{improve the model performance}. In other words, in large $n$ scenarios where SecAgg can fail (due to large communication), our protocol may enable a practitioner to still increase $n$ to obtain better performance.

We attempted to improve compression rates by compressing layers separately (per-layer) or thresholding the noisy aggregate random projections. We found neither achieved significant improvements. The former aligns with results from~\citet{mcmahan2017learning} in central DP. See Appendix~\ref{app:ssec:per-layer-threshold} for this.

\newpage
\section{Conclusion}\label{sec:conc}
In this paper, we study the optimal privacy-communication-accuracy trade-offs under distributed DP via SecAgg. We show that existing schemes are order optimal when $d \ll n^2 \varepsilon^2$ and strictly sub-optimal otherwise. To address this issue, we provide an optimal scheme that leverages sparse random projections. We also show how our scheme can be minimally modified when the client updates are sparse to further improve the trade-offs. Our extensive experiments on FL benchmark datasets demonstrate significant communication gains ($\sim$10x) relative to existing schemes. Many important questions remain open, including obtaining a fundamental characterization of the privacy-accuracy-communication trade-offs under other models of distributed DP (e.g. via a trusted third-party or in a secure enclave as in \cite{bittau2017prochlo, GKMP20-icml, anon-power, ghazi2019private, ghazi2020pure, ishai2006cryptography, BalleBGN19, balle_merged, balcer2019separating, balcer2021connecting, girgis2020shuffled, girgis2021shuffled, erlingsson2019amplification}).

\newpage

\bibliography{main.bib}

\begin{thebibliography}{62}
\providecommand{\natexlab}[1]{#1}
\providecommand{\url}[1]{\texttt{#1}}
\expandafter\ifx\csname urlstyle\endcsname\relax
  \providecommand{\doi}[1]{doi: #1}\else
  \providecommand{\doi}{doi: \begingroup \urlstyle{rm}\Url}\fi

\bibitem[Agarwal et~al.(2018)Agarwal, Suresh, Yu, Kumar, and
  McMahan]{agarwal2018cpsgd}
Agarwal, N., Suresh, A.~T., Yu, F. X.~X., Kumar, S., and McMahan, B.
\newblock cpsgd: Communication-efficient and differentially-private distributed
  sgd.
\newblock In \emph{Advances in Neural Information Processing Systems}, pp.\
  7564--7575, 2018.

\bibitem[Agarwal et~al.(2021)Agarwal, Kairouz, and Liu]{agarwal2021skellam}
Agarwal, N., Kairouz, P., and Liu, Z.
\newblock The skellam mechanism for differentially private federated learning.
\newblock \emph{Advances in Neural Information Processing Systems}, 34, 2021.

\bibitem[Aji \& Heafield(2017)Aji and Heafield]{aji2017sparse}
Aji, A.~F. and Heafield, K.
\newblock Sparse communication for distributed gradient descent.
\newblock \emph{arXiv preprint arXiv:1704.05021}, 2017.

\bibitem[Alistarh et~al.(2017)Alistarh, Grubic, Li, Tomioka, and
  Vojnovic]{alistarh17qsgd}
Alistarh, D., Grubic, D., Li, J., Tomioka, R., and Vojnovic, M.
\newblock Qsgd: Communication-efficient sgd via gradient quantization and
  encoding.
\newblock In Guyon, I., Luxburg, U.~V., Bengio, S., Wallach, H., Fergus, R.,
  Vishwanathan, S., and Garnett, R. (eds.), \emph{Advances in Neural
  Information Processing Systems}, pp.\  1709--1720. Curran Associates, Inc.,
  2017.

\bibitem[Andrew et~al.(2021)Andrew, Thakkar, McMahan, and
  Ramaswamy]{andrew2019differentially}
Andrew, G., Thakkar, O., McMahan, B., and Ramaswamy, S.
\newblock Differentially private learning with adaptive clipping.
\newblock \emph{Advances in Neural Information Processing Systems}, 34, 2021.

\bibitem[Anonymous(2022)]{anonymous2022iterative}
Anonymous.
\newblock Iterative sketching and its application to federated learning.
\newblock In \emph{Submitted to The Tenth International Conference on Learning
  Representations}, 2022.
\newblock URL \url{https://openreview.net/forum?id=U_Jog0t3fAu}.
\newblock under review.

\bibitem[Asoodeh et~al.(2020)Asoodeh, Liao, Calmon, Kosut, and
  Sankar]{asoodeh2020better}
Asoodeh, S., Liao, J., Calmon, F.~P., Kosut, O., and Sankar, L.
\newblock A better bound gives a hundred rounds: Enhanced privacy guarantees
  via f-divergences.
\newblock In \emph{2020 IEEE International Symposium on Information Theory
  (ISIT)}, pp.\  920--925. IEEE, 2020.

\bibitem[Balcer \& Cheu(2020)Balcer and Cheu]{balcer2019separating}
Balcer, V. and Cheu, A.
\newblock Separating local \& shuffled differential privacy via histograms.
\newblock In \emph{ITC}, pp.\  1:1--1:14, 2020.

\bibitem[Balcer et~al.(2021)Balcer, Cheu, Joseph, and
  Mao]{balcer2021connecting}
Balcer, V., Cheu, A., Joseph, M., and Mao, J.
\newblock Connecting robust shuffle privacy and pan-privacy.
\newblock In \emph{Proceedings of the 2021 ACM-SIAM Symposium on Discrete
  Algorithms (SODA)}, pp.\  2384--2403. SIAM, 2021.

\bibitem[Balle et~al.(2019)Balle, Bell, Gasc{\'o}n, and Nissim]{BalleBGN19}
Balle, B., Bell, J., Gasc{\'o}n, A., and Nissim, K.
\newblock The privacy blanket of the shuffle model.
\newblock In \emph{Annual International Cryptology Conference}, pp.\  638--667.
  Springer, 2019.

\bibitem[Balle et~al.(2020)Balle, Bell, Gasc{\'o}n, and Nissim]{balle_merged}
Balle, B., Bell, J., Gasc{\'o}n, A., and Nissim, K.
\newblock Private summation in the multi-message shuffle model.
\newblock In \emph{Proceedings of the 2020 ACM SIGSAC Conference on Computer
  and Communications Security}, pp.\  657--676, 2020.

\bibitem[Barnes et~al.(2020)Barnes, Han, and Ozgur]{barnes2019lower}
Barnes, L.~P., Han, Y., and Ozgur, A.
\newblock Lower bounds for learning distributions under communication
  constraints via fisher information.
\newblock \emph{Journal of Machine Learning Research}, 21\penalty0
  (236):\penalty0 1--30, 2020.

\bibitem[Bassily et~al.(2014)Bassily, Smith, and Thakurta]{bassily2014private}
Bassily, R., Smith, A., and Thakurta, A.
\newblock Private empirical risk minimization: Efficient algorithms and tight
  error bounds.
\newblock In \emph{2014 IEEE 55th Annual Symposium on Foundations of Computer
  Science}, pp.\  464--473. IEEE, 2014.

\bibitem[Bell et~al.(2020)Bell, Bonawitz, Gasc{\'o}n, Lepoint, and
  Raykova]{bell2020secure}
Bell, J.~H., Bonawitz, K.~A., Gasc{\'o}n, A., Lepoint, T., and Raykova, M.
\newblock Secure single-server aggregation with (poly) logarithmic overhead.
\newblock In \emph{Proceedings of the 2020 ACM SIGSAC Conference on Computer
  and Communications Security}, pp.\  1253--1269, 2020.

\bibitem[Bernstein et~al.(2018)Bernstein, Wang, Azizzadenesheli, and
  Anandkumar]{bernstein2018signsgd}
Bernstein, J., Wang, Y.-X., Azizzadenesheli, K., and Anandkumar, A.
\newblock signsgd: Compressed optimisation for non-convex problems.
\newblock In \emph{International Conference on Machine Learning}, pp.\
  560--569. PMLR, 2018.

\bibitem[Bittau et~al.(2017)Bittau, Erlingsson, Maniatis, Mironov, Raghunathan,
  Lie, Rudominer, Kode, Tinnes, and Seefeld]{bittau2017prochlo}
Bittau, A., Erlingsson, {\'U}., Maniatis, P., Mironov, I., Raghunathan, A.,
  Lie, D., Rudominer, M., Kode, U., Tinnes, J., and Seefeld, B.
\newblock Prochlo: Strong privacy for analytics in the crowd.
\newblock In \emph{Proceedings of the 26th Symposium on Operating Systems
  Principles}, pp.\  441--459, 2017.

\bibitem[Bonawitz et~al.(2016)Bonawitz, Ivanov, Kreuter, Marcedone, McMahan,
  Patel, Ramage, Segal, and Seth]{bonawitz2016practical}
Bonawitz, K., Ivanov, V., Kreuter, B., Marcedone, A., McMahan, H.~B., Patel,
  S., Ramage, D., Segal, A., and Seth, K.
\newblock Practical secure aggregation for federated learning on user-held
  data.
\newblock \emph{arXiv preprint arXiv:1611.04482}, 2016.

\bibitem[Bonawitz et~al.(2019)Bonawitz, Eichner, Grieskamp, Huba, Ingerman,
  Ivanov, Kiddon, Kone{\v{c}}n{\`y}, Mazzocchi, McMahan,
  et~al.]{bonawitz2019towards}
Bonawitz, K., Eichner, H., Grieskamp, W., Huba, D., Ingerman, A., Ivanov, V.,
  Kiddon, C., Kone{\v{c}}n{\`y}, J., Mazzocchi, S., McMahan, H.~B., et~al.
\newblock Towards federated learning at scale: System design.
\newblock \emph{arXiv preprint arXiv:1902.01046}, 2019.

\bibitem[Bun \& Steinke(2016)Bun and Steinke]{bun2016concentrated}
Bun, M. and Steinke, T.
\newblock Concentrated differential privacy: Simplifications, extensions, and
  lower bounds.
\newblock In \emph{Theory of Cryptography Conference}, pp.\  635--658.
  Springer, 2016.

\bibitem[Caldas et~al.(2018)Caldas, Duddu, Wu, Li, Kone{\v{c}}n{\`y}, McMahan,
  Smith, and Talwalkar]{caldas2018leaf}
Caldas, S., Duddu, S. M.~K., Wu, P., Li, T., Kone{\v{c}}n{\`y}, J., McMahan,
  H.~B., Smith, V., and Talwalkar, A.
\newblock Leaf: A benchmark for federated settings.
\newblock \emph{arXiv preprint arXiv:1812.01097}, 2018.

\bibitem[Canonne et~al.(2020)Canonne, Kamath, and Steinke]{canonne2020discrete}
Canonne, C.~L., Kamath, G., and Steinke, T.
\newblock The discrete gaussian for differential privacy.
\newblock \emph{arXiv preprint arXiv:2004.00010}, 2020.

\bibitem[Carlini et~al.(2019)Carlini, Liu, Erlingsson, Kos, and
  Song]{carlini2019secret}
Carlini, N., Liu, C., Erlingsson, {\'U}., Kos, J., and Song, D.
\newblock The secret sharer: Evaluating and testing unintended memorization in
  neural networks.
\newblock In \emph{28th $\{$USENIX$\}$ Security Symposium ($\{$USENIX$\}$
  Security 19)}, pp.\  267--284, 2019.

\bibitem[Chen et~al.(2020)Chen, Kairouz, and Ozgur]{chen2020breaking}
Chen, W.-N., Kairouz, P., and Ozgur, A.
\newblock Breaking the communication-privacy-accuracy trilemma.
\newblock \emph{Advances in Neural Information Processing Systems},
  33:\penalty0 3312--3324, 2020.

\bibitem[Duchi et~al.(2013)Duchi, Jordan, and Wainwright]{duchi2013local}
Duchi, J.~C., Jordan, M.~I., and Wainwright, M.~J.
\newblock Local privacy and statistical minimax rates.
\newblock In \emph{2013 IEEE 54th Annual Symposium on Foundations of Computer
  Science}, pp.\  429--438. IEEE, 2013.

\bibitem[Dwork et~al.(2006{\natexlab{a}})Dwork, Kenthapadi, McSherry, Mironov,
  and Naor]{dwork2006our}
Dwork, C., Kenthapadi, K., McSherry, F., Mironov, I., and Naor, M.
\newblock Our data, ourselves: Privacy via distributed noise generation.
\newblock In \emph{Annual International Conference on the Theory and
  Applications of Cryptographic Techniques}, pp.\  486--503. Springer,
  2006{\natexlab{a}}.

\bibitem[Dwork et~al.(2006{\natexlab{b}})Dwork, McSherry, Nissim, and
  Smith]{dwork2006calibrating}
Dwork, C., McSherry, F., Nissim, K., and Smith, A.
\newblock Calibrating noise to sensitivity in private data analysis.
\newblock In \emph{Theory of cryptography conference}, pp.\  265--284.
  Springer, 2006{\natexlab{b}}.

\bibitem[Erlingsson et~al.(2019)Erlingsson, Feldman, Mironov, Raghunathan,
  Talwar, and Thakurta]{erlingsson2019amplification}
Erlingsson, {\'U}., Feldman, V., Mironov, I., Raghunathan, A., Talwar, K., and
  Thakurta, A.
\newblock Amplification by shuffling: From local to central differential
  privacy via anonymity.
\newblock In \emph{Proceedings of the Thirtieth Annual ACM-SIAM Symposium on
  Discrete Algorithms}, pp.\  2468--2479. SIAM, 2019.

\bibitem[Evfimievski et~al.(2004)Evfimievski, Srikant, Agrawal, and
  Gehrke]{evfimievski2004privacy}
Evfimievski, A., Srikant, R., Agrawal, R., and Gehrke, J.
\newblock Privacy preserving mining of association rules.
\newblock \emph{Information Systems}, 29\penalty0 (4):\penalty0 343--364, 2004.

\bibitem[Geyer et~al.(2017)Geyer, Klein, and Nabi]{geyer2017differentially}
Geyer, R.~C., Klein, T., and Nabi, M.
\newblock Differentially private federated learning: A client level
  perspective.
\newblock \emph{arXiv preprint arXiv:1712.07557}, 2017.

\bibitem[Ghazi et~al.(2020{\natexlab{a}})Ghazi, Golowich, Kumar, Manurangsi,
  Pagh, and Velingker]{ghazi2020pure}
Ghazi, B., Golowich, N., Kumar, R., Manurangsi, P., Pagh, R., and Velingker, A.
\newblock Pure differentially private summation from anonymous messages.
\newblock In \emph{1st Conference on Information-Theoretic Cryptography (ITC
  2020)}. Schloss Dagstuhl-Leibniz-Zentrum f{\"u}r Informatik,
  2020{\natexlab{a}}.

\bibitem[Ghazi et~al.(2020{\natexlab{b}})Ghazi, Kumar, Manurangsi, and
  Pagh]{GKMP20-icml}
Ghazi, B., Kumar, R., Manurangsi, P., and Pagh, R.
\newblock Private counting from anonymous messages: Near-optimal accuracy with
  vanishing communication overhead.
\newblock In \emph{ICML}, pp.\  3505--3514, 2020{\natexlab{b}}.

\bibitem[Ghazi et~al.(2020{\natexlab{c}})Ghazi, Manurangsi, Pagh, and
  Velingker]{ghazi2019private}
Ghazi, B., Manurangsi, P., Pagh, R., and Velingker, A.
\newblock Private aggregation from fewer anonymous messages.
\newblock In \emph{Eurocrypt}, 2020{\natexlab{c}}.

\bibitem[Ghazi et~al.(2021)Ghazi, Golowich, Kumar, Pagh, and
  Velingker]{anon-power}
Ghazi, B., Golowich, N., Kumar, R., Pagh, R., and Velingker, A.
\newblock On the power of multiple anonymous messages.
\newblock In \emph{Eurocrypt}, 2021.
\newblock To appear.

\bibitem[Girgis et~al.(2021{\natexlab{a}})Girgis, Data, Diggavi, Kairouz, and
  Suresh]{girgis2021shuffled}
Girgis, A., Data, D., Diggavi, S., Kairouz, P., and Suresh, A.~T.
\newblock Shuffled model of differential privacy in federated learning.
\newblock In \emph{International Conference on Artificial Intelligence and
  Statistics}, pp.\  2521--2529. PMLR, 2021{\natexlab{a}}.

\bibitem[Girgis et~al.(2021{\natexlab{b}})Girgis, Data, Diggavi, Kairouz, and
  Suresh]{girgis2020shuffled}
Girgis, A.~M., Data, D., Diggavi, S., Kairouz, P., and Suresh, A.~T.
\newblock Shuffled model of federated learning: Privacy, accuracy and
  communication trade-offs.
\newblock \emph{IEEE Journal on Selected Areas in Information Theory},
  2\penalty0 (1):\penalty0 464--478, 2021{\natexlab{b}}.
\newblock \doi{10.1109/JSAIT.2021.3056102}.

\bibitem[Haddadpour et~al.(2020)Haddadpour, Karimi, Li, and
  Li]{haddadpour2020fedsketch}
Haddadpour, F., Karimi, B., Li, P., and Li, X.
\newblock Fedsketch: Communication-efficient and private federated learning via
  sketching.
\newblock \emph{arXiv preprint arXiv:2008.04975}, 2020.

\bibitem[Havasi et~al.(2018)Havasi, Peharz, and
  Hern{\'a}ndez-Lobato]{havasi2018minimal}
Havasi, M., Peharz, R., and Hern{\'a}ndez-Lobato, J.~M.
\newblock Minimal random code learning: Getting bits back from compressed model
  parameters.
\newblock \emph{arXiv preprint arXiv:1810.00440}, 2018.

\bibitem[Ishai et~al.(2006)Ishai, Kushilevitz, Ostrovsky, and
  Sahai]{ishai2006cryptography}
Ishai, Y., Kushilevitz, E., Ostrovsky, R., and Sahai, A.
\newblock Cryptography from anonymity.
\newblock In \emph{2006 47th Annual IEEE Symposium on Foundations of Computer
  Science (FOCS'06)}, pp.\  239--248. IEEE, 2006.

\bibitem[Kairouz et~al.(2016)Kairouz, Bonawitz, and Ramage]{kairouz16}
Kairouz, P., Bonawitz, K., and Ramage, D.
\newblock Discrete distribution estimation under local privacy.
\newblock In \emph{Proceedings of The 33rd International Conference on Machine
  Learning}, volume~48, pp.\  2436--2444, New York, New York, USA, 20--22 Jun
  2016.

\bibitem[Kairouz et~al.(2019)Kairouz, McMahan, Avent, Bellet, Bennis, Bhagoji,
  Bonawitz, Charles, Cormode, Cummings, et~al.]{kairouz2019advances}
Kairouz, P., McMahan, H.~B., Avent, B., Bellet, A., Bennis, M., Bhagoji, A.~N.,
  Bonawitz, K., Charles, Z., Cormode, G., Cummings, R., et~al.
\newblock Advances and open problems in federated learning.
\newblock \emph{arXiv preprint arXiv:1912.04977}, 2019.

\bibitem[Kairouz et~al.(2021{\natexlab{a}})Kairouz, Liu, and
  Steinke]{kairouz2021distributed}
Kairouz, P., Liu, Z., and Steinke, T.
\newblock The distributed discrete gaussian mechanism for federated learning
  with secure aggregation.
\newblock In \emph{International Conference on Machine Learning}, pp.\
  5201--5212. PMLR, 2021{\natexlab{a}}.

\bibitem[Kairouz et~al.(2021{\natexlab{b}})Kairouz, McMahan, Avent, Bellet,
  Bennis, Bhagoji, Bonawitz, Charles, Cormode, Cummings, D’Oliveira, Eichner,
  Rouayheb, Evans, Gardner, Garrett, Gascón, Ghazi, Gibbons, Gruteser,
  Harchaoui, He, He, Huo, Hutchinson, Hsu, Jaggi, Javidi, Joshi, Khodak,
  Konecný, Korolova, Koushanfar, Koyejo, Lepoint, Liu, Mittal, Mohri, Nock,
  Özgür, Pagh, Qi, Ramage, Raskar, Raykova, Song, Song, Stich, Sun, Suresh,
  Tramèr, Vepakomma, Wang, Xiong, Xu, Yang, Yu, Yu, and Zhao]{aopfl}
Kairouz, P., McMahan, H.~B., Avent, B., Bellet, A., Bennis, M., Bhagoji, A.~N.,
  Bonawitz, K., Charles, Z., Cormode, G., Cummings, R., D’Oliveira, R. G.~L.,
  Eichner, H., Rouayheb, S.~E., Evans, D., Gardner, J., Garrett, Z., Gascón,
  A., Ghazi, B., Gibbons, P.~B., Gruteser, M., Harchaoui, Z., He, C., He, L.,
  Huo, Z., Hutchinson, B., Hsu, J., Jaggi, M., Javidi, T., Joshi, G., Khodak,
  M., Konecný, J., Korolova, A., Koushanfar, F., Koyejo, S., Lepoint, T., Liu,
  Y., Mittal, P., Mohri, M., Nock, R., Özgür, A., Pagh, R., Qi, H., Ramage,
  D., Raskar, R., Raykova, M., Song, D., Song, W., Stich, S.~U., Sun, Z.,
  Suresh, A.~T., Tramèr, F., Vepakomma, P., Wang, J., Xiong, L., Xu, Z., Yang,
  Q., Yu, F.~X., Yu, H., and Zhao, S.
\newblock Advances and open problems in federated learning.
\newblock \emph{Foundations and Trends® in Machine Learning}, 14\penalty0
  (1–2):\penalty0 1--210, 2021{\natexlab{b}}.
\newblock ISSN 1935-8237.
\newblock \doi{10.1561/2200000083}.
\newblock URL \url{http://dx.doi.org/10.1561/2200000083}.

\bibitem[Kamath \& Ullman(2020)Kamath and Ullman]{kamath2020primer}
Kamath, G. and Ullman, J.
\newblock A primer on private statistics.
\newblock \emph{arXiv preprint arXiv:2005.00010}, 2020.

\bibitem[Kane \& Nelson(2014)Kane and Nelson]{kane2014sparser}
Kane, D.~M. and Nelson, J.
\newblock Sparser johnson-lindenstrauss transforms.
\newblock \emph{Journal of the ACM (JACM)}, 61\penalty0 (1):\penalty0 1--23,
  2014.

\bibitem[Kasiviswanathan et~al.(2011)Kasiviswanathan, Lee, Nissim,
  Raskhodnikova, and Smith]{kasiviswanathan2011can}
Kasiviswanathan, S.~P., Lee, H.~K., Nissim, K., Raskhodnikova, S., and Smith,
  A.
\newblock What can we learn privately?
\newblock \emph{SIAM Journal on Computing}, 40\penalty0 (3):\penalty0 793--826,
  2011.

\bibitem[Lin et~al.(2017)Lin, Han, Mao, Wang, and Dally]{lin2017deep}
Lin, Y., Han, S., Mao, H., Wang, Y., and Dally, W.~J.
\newblock Deep gradient compression: Reducing the communication bandwidth for
  distributed training.
\newblock \emph{arXiv preprint arXiv:1712.01887}, 2017.

\bibitem[McMahan et~al.(2017{\natexlab{a}})McMahan, Moore, Ramage, Hampson, and
  y~Arcas]{mcmahan2017communication}
McMahan, B., Moore, E., Ramage, D., Hampson, S., and y~Arcas, B.~A.
\newblock Communication-efficient learning of deep networks from decentralized
  data.
\newblock In \emph{Artificial intelligence and statistics}, pp.\  1273--1282.
  PMLR, 2017{\natexlab{a}}.

\bibitem[McMahan et~al.(2017{\natexlab{b}})McMahan, Ramage, Talwar, and
  Zhang]{mcmahan2017learning}
McMahan, H.~B., Ramage, D., Talwar, K., and Zhang, L.
\newblock Learning differentially private recurrent language models.
\newblock \emph{arXiv preprint arXiv:1710.06963}, 2017{\natexlab{b}}.

\bibitem[Melis et~al.(2019)Melis, Song, De~Cristofaro, and
  Shmatikov]{melis2019exploiting}
Melis, L., Song, C., De~Cristofaro, E., and Shmatikov, V.
\newblock Exploiting unintended feature leakage in collaborative learning.
\newblock In \emph{2019 IEEE Symposium on Security and Privacy (SP)}, pp.\
  691--706. IEEE, 2019.

\bibitem[Mironov(2017)]{mironov2017renyi}
Mironov, I.
\newblock R{\'e}nyi differential privacy.
\newblock In \emph{2017 IEEE 30th Computer Security Foundations Symposium
  (CSF)}, pp.\  263--275. IEEE, 2017.

\bibitem[Oktay et~al.(2019)Oktay, Ball{\'e}, Singh, and
  Shrivastava]{oktay2019scalable}
Oktay, D., Ball{\'e}, J., Singh, S., and Shrivastava, A.
\newblock Scalable model compression by entropy penalized reparameterization.
\newblock \emph{arXiv preprint arXiv:1906.06624}, 2019.

\bibitem[Polyak(1964)]{polyak1964some}
Polyak, B.~T.
\newblock Some methods of speeding up the convergence of iteration methods.
\newblock \emph{Ussr computational mathematics and mathematical physics},
  4\penalty0 (5):\penalty0 1--17, 1964.

\bibitem[Raskutti et~al.(2010)Raskutti, Wainwright, and
  Yu]{raskutti2010restricted}
Raskutti, G., Wainwright, M.~J., and Yu, B.
\newblock Restricted eigenvalue properties for correlated gaussian designs.
\newblock \emph{The Journal of Machine Learning Research}, 11:\penalty0
  2241--2259, 2010.

\bibitem[Rothchild et~al.(2020)Rothchild, Panda, Ullah, Ivkin, Stoica,
  Braverman, Gonzalez, and Arora]{rothchild2020fetchsgd}
Rothchild, D., Panda, A., Ullah, E., Ivkin, N., Stoica, I., Braverman, V.,
  Gonzalez, J., and Arora, R.
\newblock Fetchsgd: Communication-efficient federated learning with sketching.
\newblock In \emph{International Conference on Machine Learning}, pp.\
  8253--8265. PMLR, 2020.

\bibitem[Shokri et~al.(2017)Shokri, Stronati, Song, and
  Shmatikov]{shokri2017membership}
Shokri, R., Stronati, M., Song, C., and Shmatikov, V.
\newblock Membership inference attacks against machine learning models.
\newblock In \emph{2017 IEEE Symposium on Security and Privacy (SP)}, pp.\
  3--18. IEEE, 2017.

\bibitem[Song \& Shmatikov(2019)Song and Shmatikov]{song2019auditing}
Song, C. and Shmatikov, V.
\newblock Auditing data provenance in text-generation models.
\newblock In \emph{Proceedings of the 25th ACM SIGKDD International Conference
  on Knowledge Discovery \& Data Mining}, pp.\  196--206, 2019.

\bibitem[Song et~al.(2013)Song, Chaudhuri, and Sarwate]{song2013stochastic}
Song, S., Chaudhuri, K., and Sarwate, A.~D.
\newblock Stochastic gradient descent with differentially private updates.
\newblock In \emph{2013 IEEE Global Conference on Signal and Information
  Processing}, pp.\  245--248. IEEE, 2013.

\bibitem[Suresh et~al.(2017)Suresh, Yu, Kumar, and McMahan]{an2016distributed}
Suresh, A.~T., Yu, F.~X., Kumar, S., and McMahan, H.~B.
\newblock Distributed mean estimation with limited communication.
\newblock In \emph{Proceedings of the 34th International Conference on Machine
  Learning - Volume 70}, ICML’17, pp.\  3329–3337. JMLR.org, 2017.

\bibitem[Tibshirani(1996)]{tibshirani1996regression}
Tibshirani, R.
\newblock Regression shrinkage and selection via the lasso.
\newblock \emph{Journal of the Royal Statistical Society: Series B
  (Methodological)}, 58\penalty0 (1):\penalty0 267--288, 1996.

\bibitem[Wainwright(2019)]{wainwright2019high}
Wainwright, M.~J.
\newblock \emph{High-dimensional statistics: A non-asymptotic viewpoint},
  volume~48.
\newblock Cambridge University Press, 2019.

\bibitem[Wangni et~al.(2017)Wangni, Wang, Liu, and Zhang]{wangni2017gradient}
Wangni, J., Wang, J., Liu, J., and Zhang, T.
\newblock Gradient sparsification for communication-efficient distributed
  optimization.
\newblock \emph{arXiv preprint arXiv:1710.09854}, 2017.

\bibitem[Warner(1965)]{warner1965randomized}
Warner, S.~L.
\newblock Randomized response: A survey technique for eliminating evasive
  answer bias.
\newblock \emph{Journal of the American Statistical Association}, 60\penalty0
  (309):\penalty0 63--69, 1965.

\end{thebibliography}
\bibliographystyle{icml2022}

\newpage
\appendix
\onecolumn
\section{Additional Figures}\label{app:figures}

\begin{figure}[h]
\centering
\subfloat[][\textbf{A large model, compressed to the same latent dimension as a small model, outperforms it.}]{\includegraphics[width=.48\textwidth]{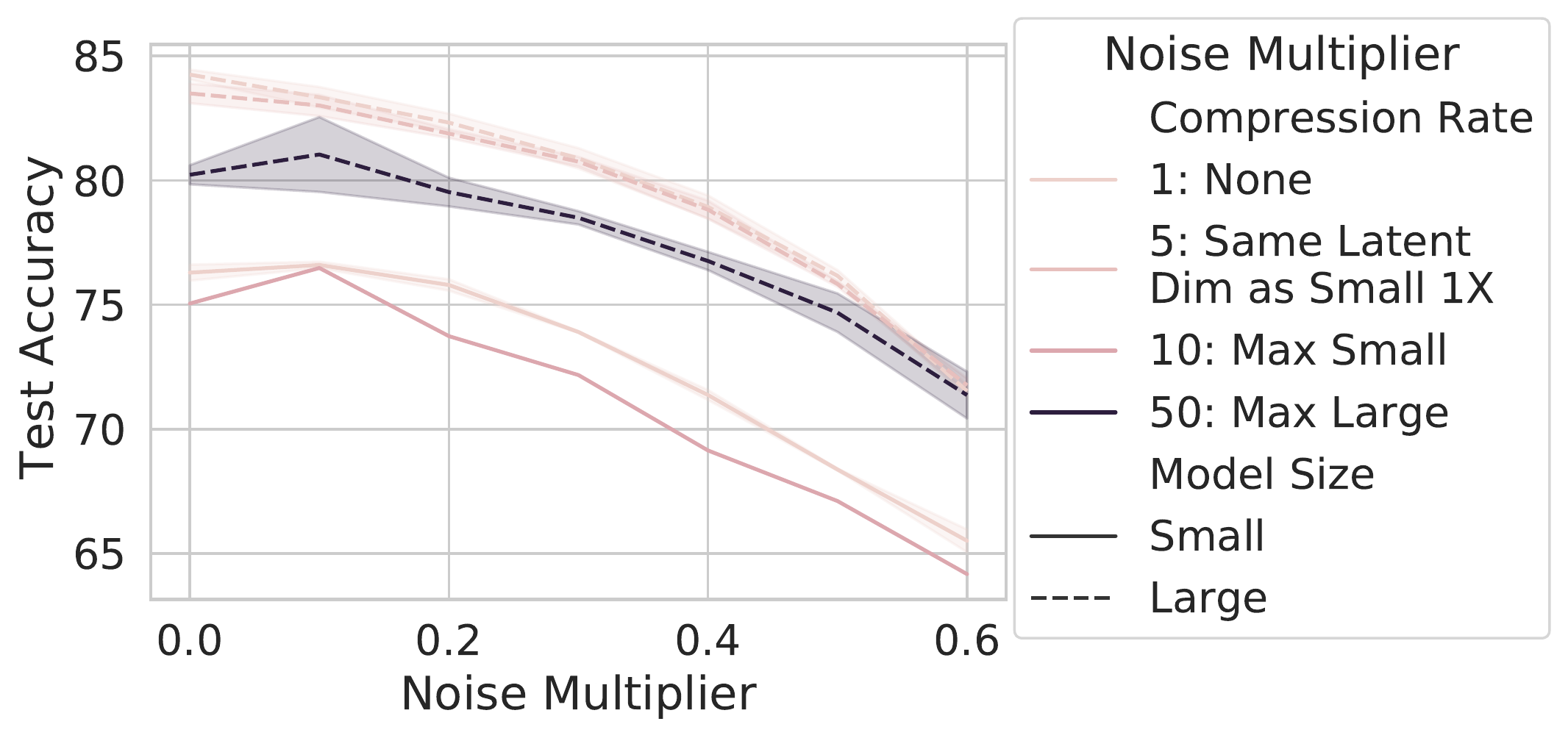}\label{fig:small-a}}\hspace{4mm}
\subfloat[][\textbf{Impact of the number of parameters on the \puc tradeoff.}]{\includegraphics[width=.48\textwidth]{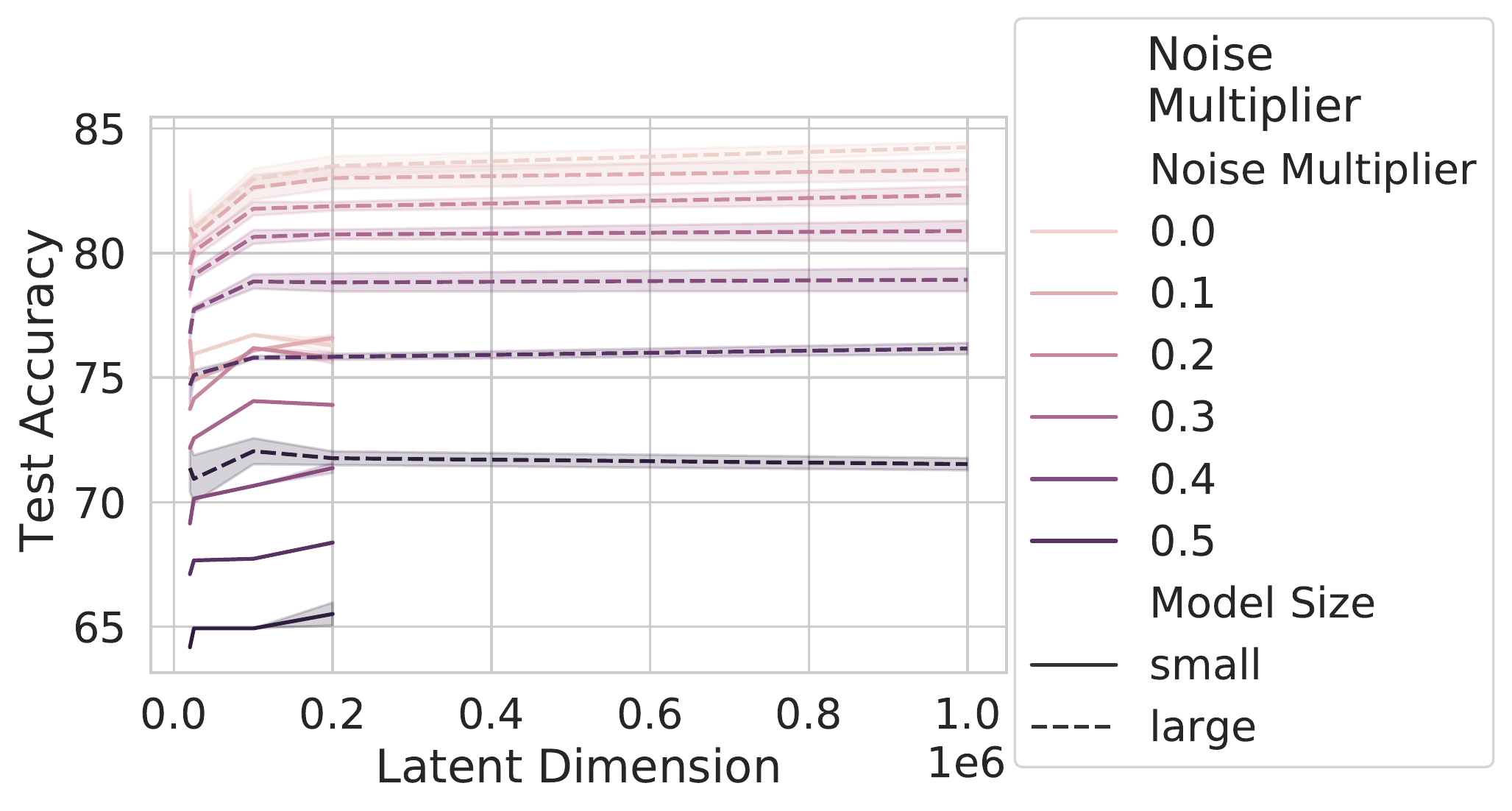}\label{fig:small-b}}

\caption{\textbf{Large models with compression outperform small models.}}
\label{fig:small}
\end{figure}

\begin{figure}[h]
    \centering
    \captionsetup[subfigure]{width=0.95\linewidth}
    \begin{subfigure}[t]{0.495\linewidth}
        \includegraphics[width=\linewidth]{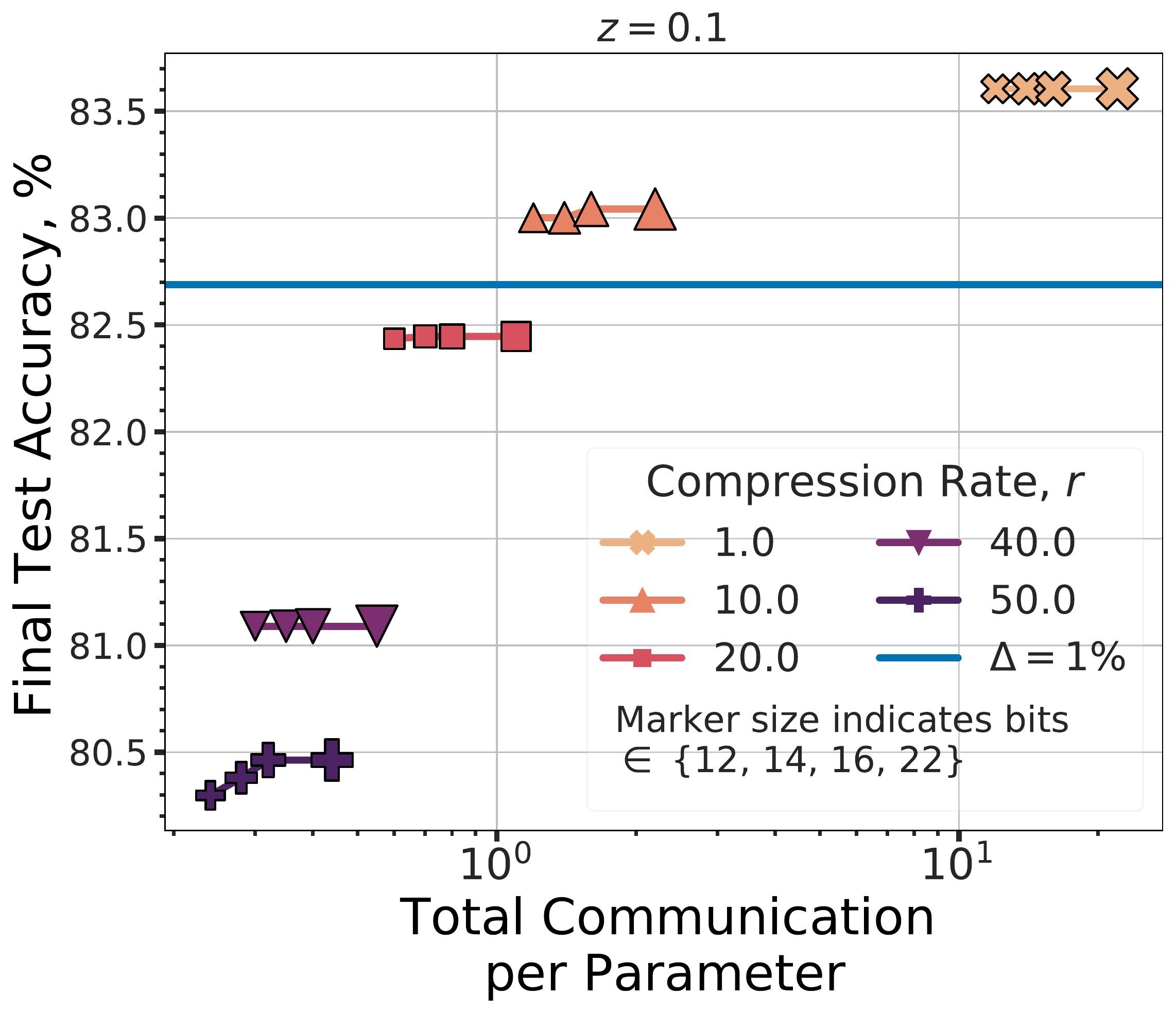}
        \vspace{-5mm}
        \caption{\textbf{Lowest communication of $1.2$ bits per parameter} at $z=0.1$ with $b=12,r=10$x.}
        \label{fig:emnist-q-b-0.1-short}
    \end{subfigure}
    \begin{subfigure}[t]{0.495\linewidth}
        \includegraphics[width=\linewidth]{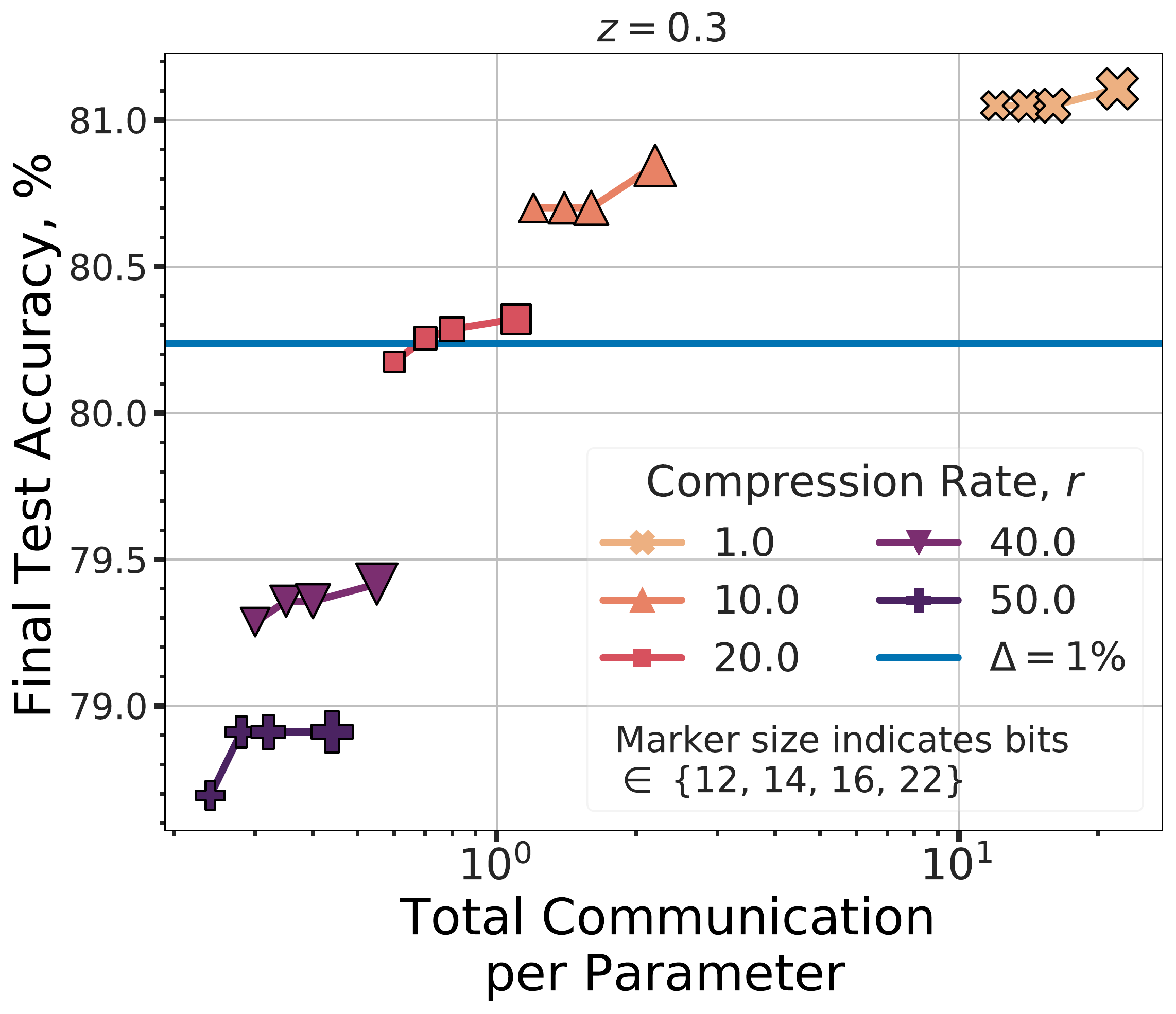}
        \vspace{-5mm}
        \caption{\textbf{Lowest communication of $0.7$ bits per parameter} at $z=0.3$ with $b=14,r=20$x.}
        \label{fig:emnist-q-b-0.3-short}
    \end{subfigure}
    \caption{\textbf{Optimizing both $r$ and $b$ can further decrease communication}. Note that sometimes at higher bitwidths we observe lower performance within statistical error (standard deviation $\approx 1$)---here, we threshold to the highest accuracy of lower bitwidths. Full results without thresholding and all $r$ are in Figure~\ref{fig:femnist-long} of Appendix~\ref{app:figures}.}
    \vspace{-1em}
    \label{fig:femnist-short}
    \vspace{-.5em}
\end{figure}

\begin{figure}[h]
    \centering
    \captionsetup[subfigure]{width=0.95\linewidth}
    \begin{subfigure}[t]{0.495\linewidth}
        \includegraphics[width=\linewidth]{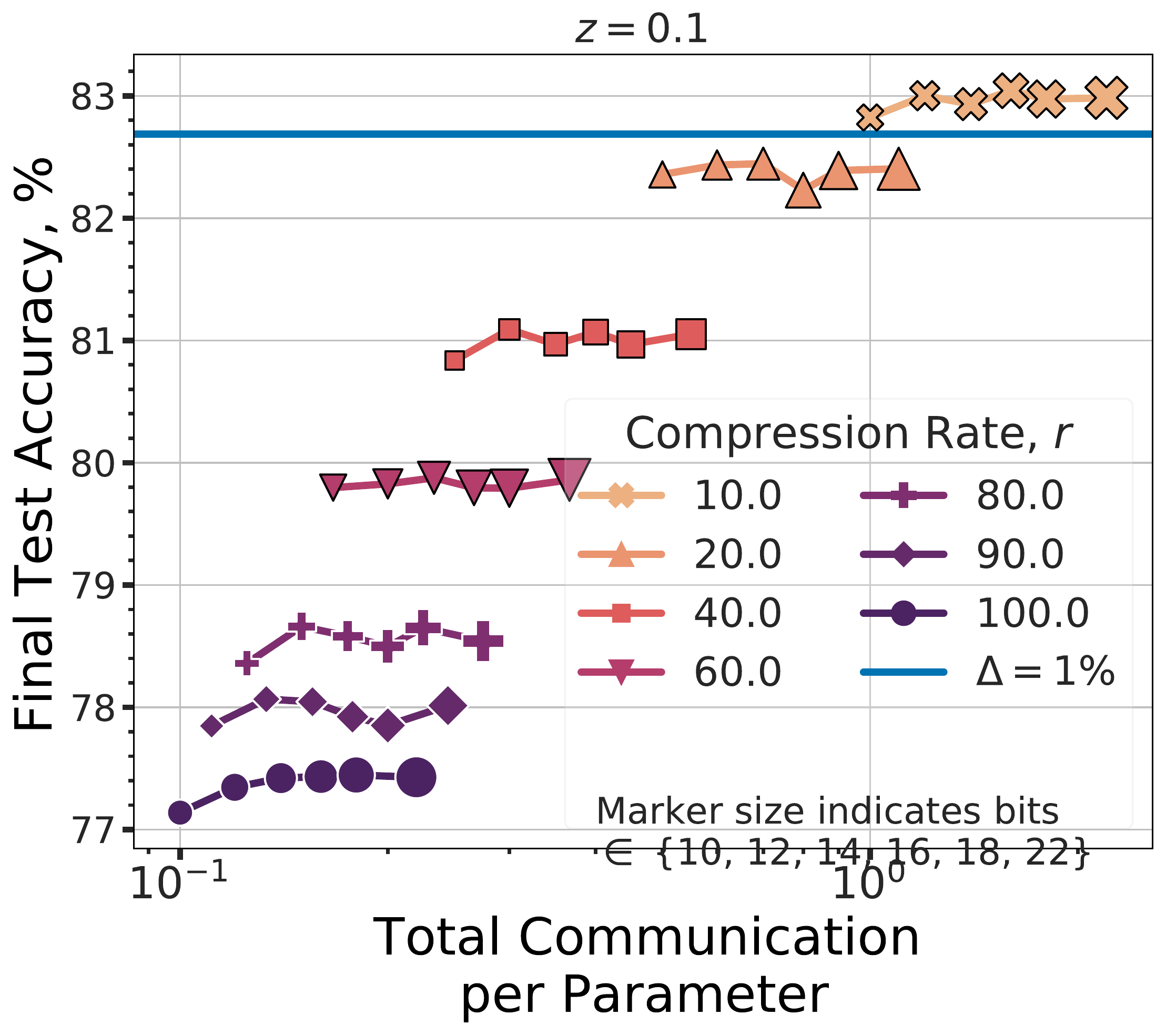}
        \vspace{-5mm}
        \caption{\textbf{Lowest communication of $1.2$ bits per parameter} at $z=0.1$ with $b=12,r=10$x.}
        \label{fig:emnist-q-b-0.1-long}
    \end{subfigure}
    \begin{subfigure}[t]{0.495\linewidth}
        \includegraphics[width=\linewidth]{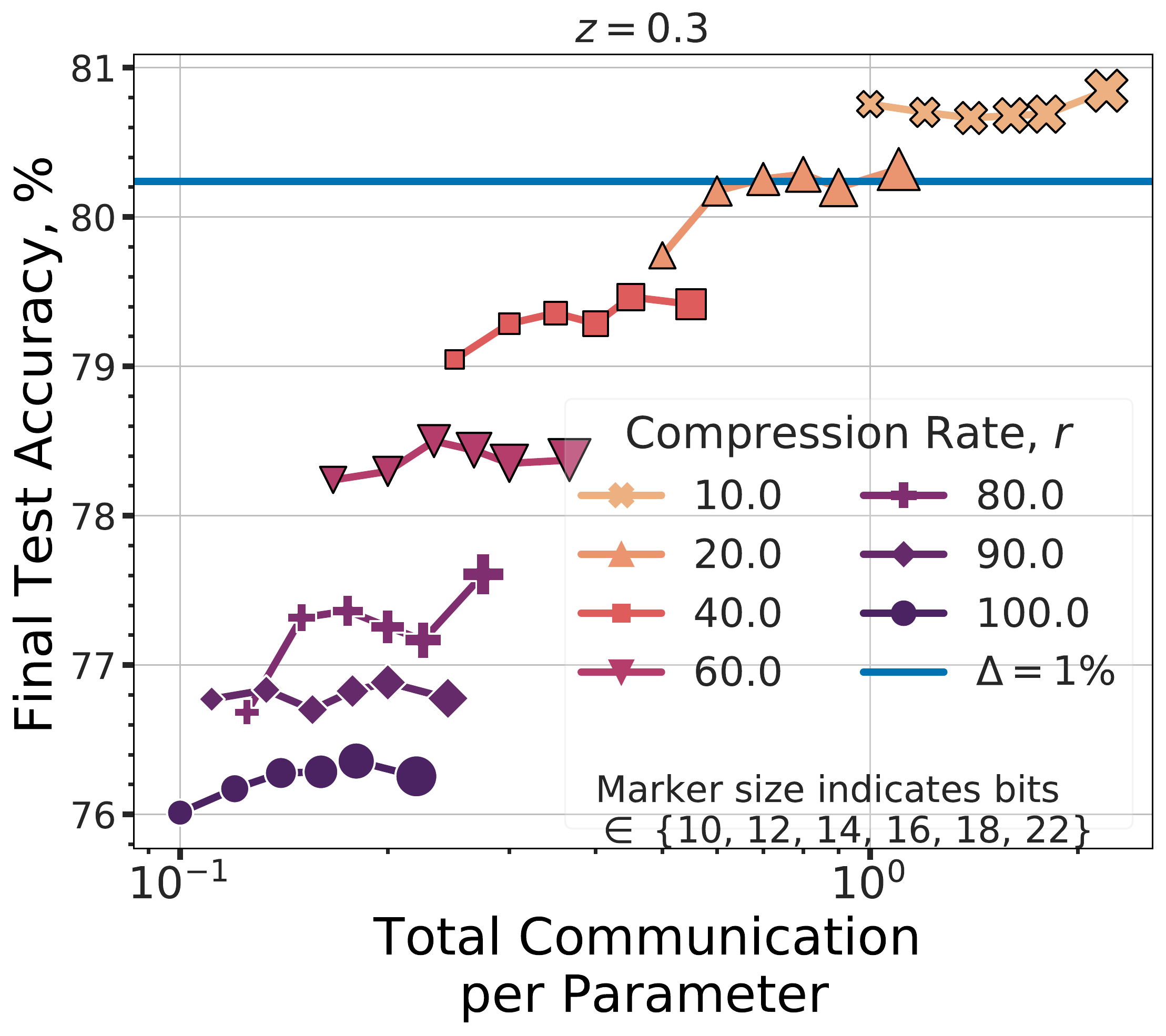}
        \vspace{-5mm}
        \caption{\textbf{Lowest communication of $0.7$ bits per parameter} at $z=0.3$ with $b=14,r=20$x.}
        \label{fig:emnist-q-b-0.3-long}
    \end{subfigure}
    \begin{subfigure}[t]{0.495\linewidth}
        \includegraphics[width=\linewidth]{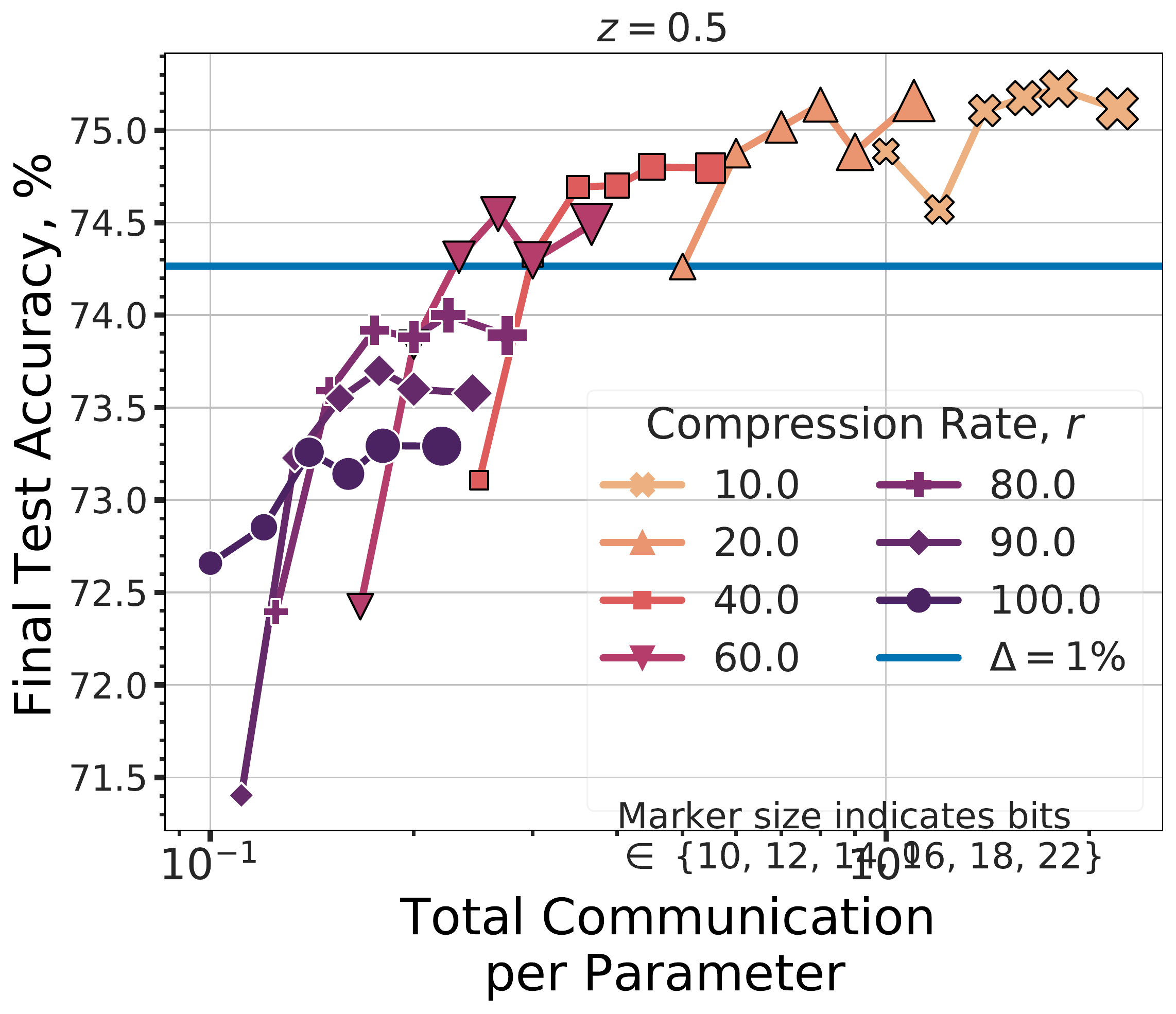}
        \vspace{-5mm}
        \caption{\textbf{Lowest communication of $0.7$ bits per parameter} at $z=0.3$ with $b=14,r=20$x.}
        \label{fig:emnist-q-b-0.5-long}
    \end{subfigure}
    \caption{\textbf{Optimizing both $r$ and $b$ can further decrease communication}. Full results for Figure~\ref{fig:quant-vary} and~\ref{fig:femnist-short}.}
    \vspace{-1em}
    \label{fig:femnist-long}
    \vspace{-.5em}
\end{figure}

\clearpage
\subsection{Impact of cohort size}\label{app:ssec:n}
\begin{figure}[h]
    \centering
    \includegraphics[width=0.6\linewidth]{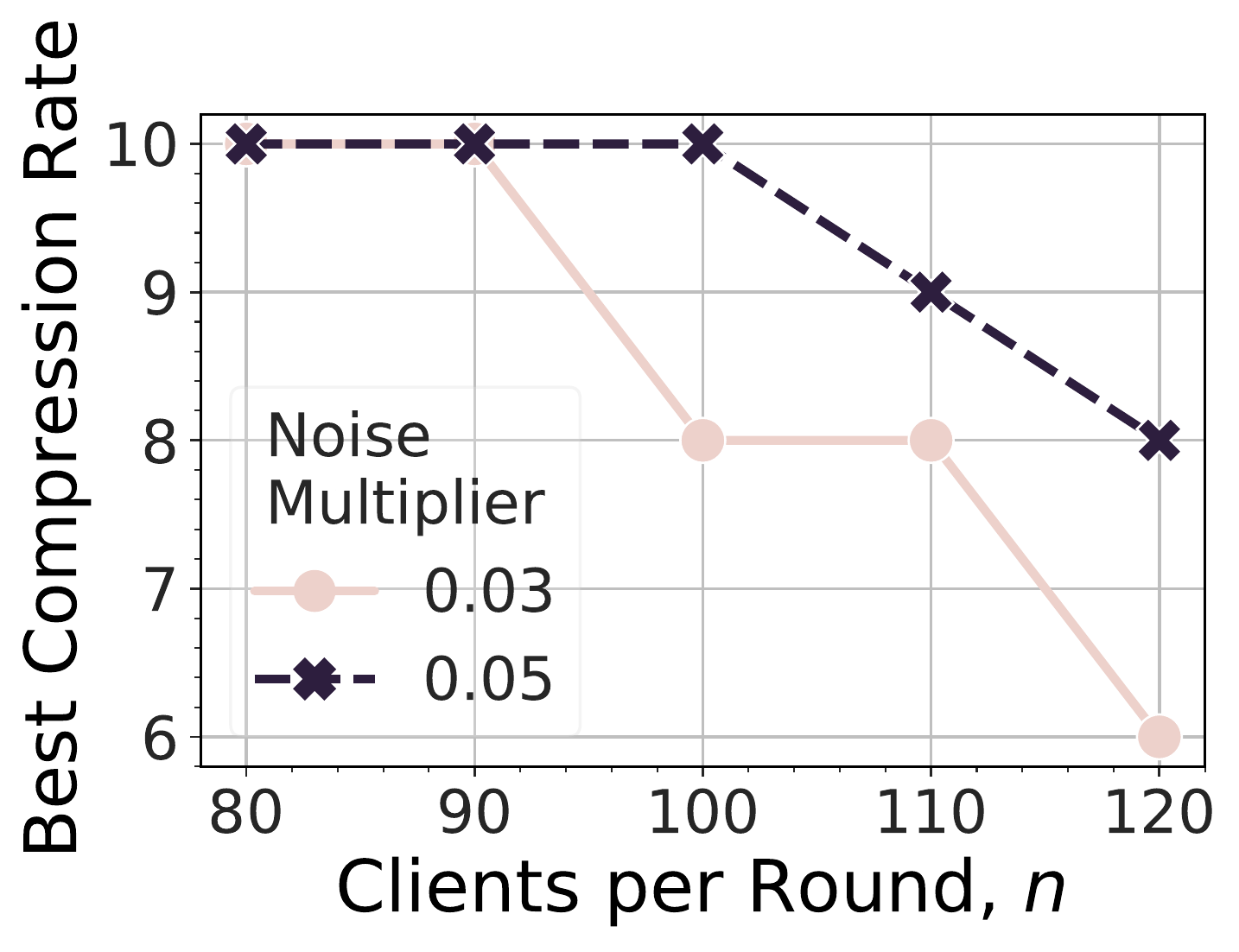} %
    \caption{\textbf{Increasing $\mathbf{n}$ leads to an increase in communication} shown by the decreasing compression rates. Higher noise multipliers can still attain higher compression. Results for SONWP.}
    \label{fig:n-sonwp}
    \begin{tabular}{c | c || c | c }
         \makecell{Noise\\Multiplier, $z$} & \makecell{Number of\\Clients, $n$} & \makecell{Compression\\Rate, $r$} & \makecell{Final Test\\Performance, \%}  \\
         \hline\hline
         \multirow{2}{*}{0.1} & 100 & 1 & $83.05\pm0.44$ \\
         & 1000 & 10 & $82.95\pm0.40$ \\
         \hline
         \multirow{2}{*}{0.3} & 100 & 1 & $80.61\pm0.46$ \\
         & 1000 & 40 & $80.78\pm0.29$ \\
         \hline
         \multirow{2}{*}{0.5} & 100 & 1 & $75.34\pm0.49$ \\
         & 1000 & 50 & $80.13\pm0.22$ \\
    \end{tabular}%
    \caption{\textbf{With $\mathbf{z}$ sufficiently large, increasing $\mathbf{n}=100\to1000$ can attain higher model performance even for increased $r$}. In particular, to maintain the same SecAgg runtime, we require $r\geq 15$ for this setting to increase $n=100\to1000$. We observe that $z\geq 0.3$ meets this requirement while achieving final models that outperform the $n=100$, $r=1$x client baseline.}
    \label{tab:n-emnist}
\end{figure}

Finally, we explore how varying the number of clients per round (or, cohort size) $n$ impacts the \puc tradeoff. This value plays several key roles in this tradeoff. First, increasing $n$ increases the sampling probability of the cohort, which increases the total privacy expenditure. However, it also tends to improve model performance---this may mean that a higher noise multiplier $z$ can be chosen so as to instead decrease the total privacy cost (this is typically the case when N is large enough, e.g., SO). In terms of communication, Theorem~\ref{thm:random_projection_upper_bound} suggests that increasing $n$ will also increase the per-client comunication. Because of the aforementioned complex tradeoffs, we (approximately) fix the privacy budget $\varepsilon$ and only perturb $n$ minimally around a nominal value of $100$. In Figure~\ref{fig:n-sonwp}, we see that dependence of communication on $n$ is observed empirically as well.

In addition to this impact on $r$, setting $n$ can also have a significant impact on the run time of SecAgg, of $O(n \log(n) d)$. 
For large values of $n$, this can entirely prevent the protocol from completing. Because a practitioner desires the most performant model, a common goal is the increase $n$ so as to obtain a tight $\varepsilon$ (due to a now higher $z$) with the least cost in performance. But, because large $n$ can crash SecAgg, this places a constraint on the maximum $n$ that can be chosen. Since our methods compress the updates ($d$ above), it is possible to still increase $n$ so long as we increase $r$ accordingly (by $\frac{n_2}{n_1}\frac{\log{n_2}}{\log{n_1}}$ where $n_2>n_1$), which maintains fixed runtime. If the resulting model at higher $n$ achieves higher performance, then we observe a net benefit from this tradeoff. For a practical privacy parameter or $z=0.6$, our results in Table~\ref{tab:n-emnist} suggest that this may be possible. Specifically, increasing $n$ from $100\to1000$ and settings $r=50x>15x$ accordingly, we observe that the final model with $n=1000$ clients achieves a nearly $5$pp gain. We observe that $z<0.3$ cannot meet these requirements.
\clearpage
\subsection{Attempting to improve compression via per-layer sketching and thresholding}\label{app:ssec:per-layer-threshold}
\begin{figure}[h]
    \centering
    \includegraphics[width=\linewidth]{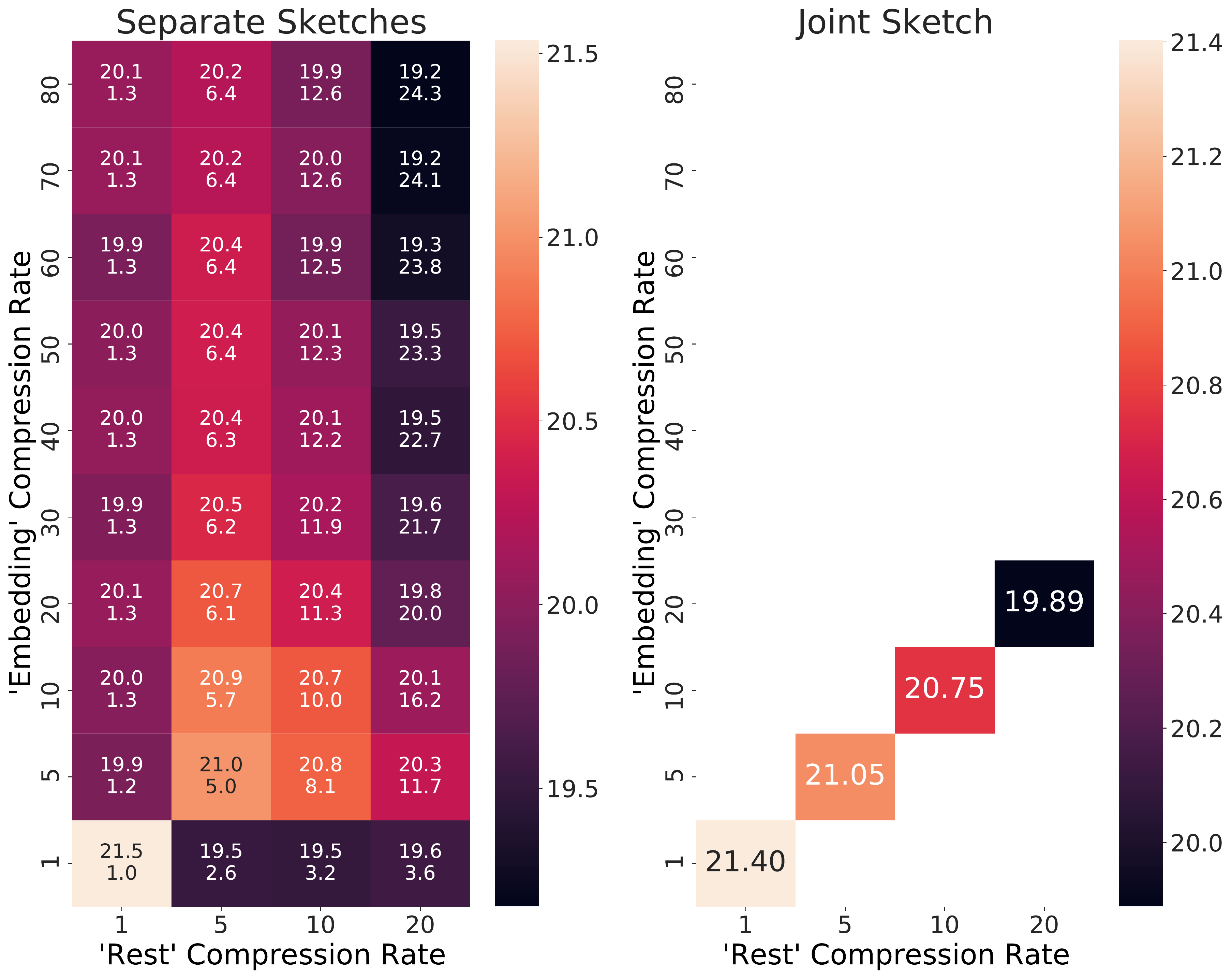}
    \caption{\textbf{Separate sketching does not significantly improve the final model performance.} Heatmap values correspond to the final model test performance followed by (newline) the total compression and are colored by the test performance. Results using the LSTM model on SONWP with $z=0.05$. We train models either by sketching the entire concatenated gradient vector or by sketching the `embedding' layer separate from the `rest' of the model.}
    \label{fig:perlayer-0.05}
\end{figure}

\begin{figure}[h]
    \centering
    \includegraphics[width=\linewidth]{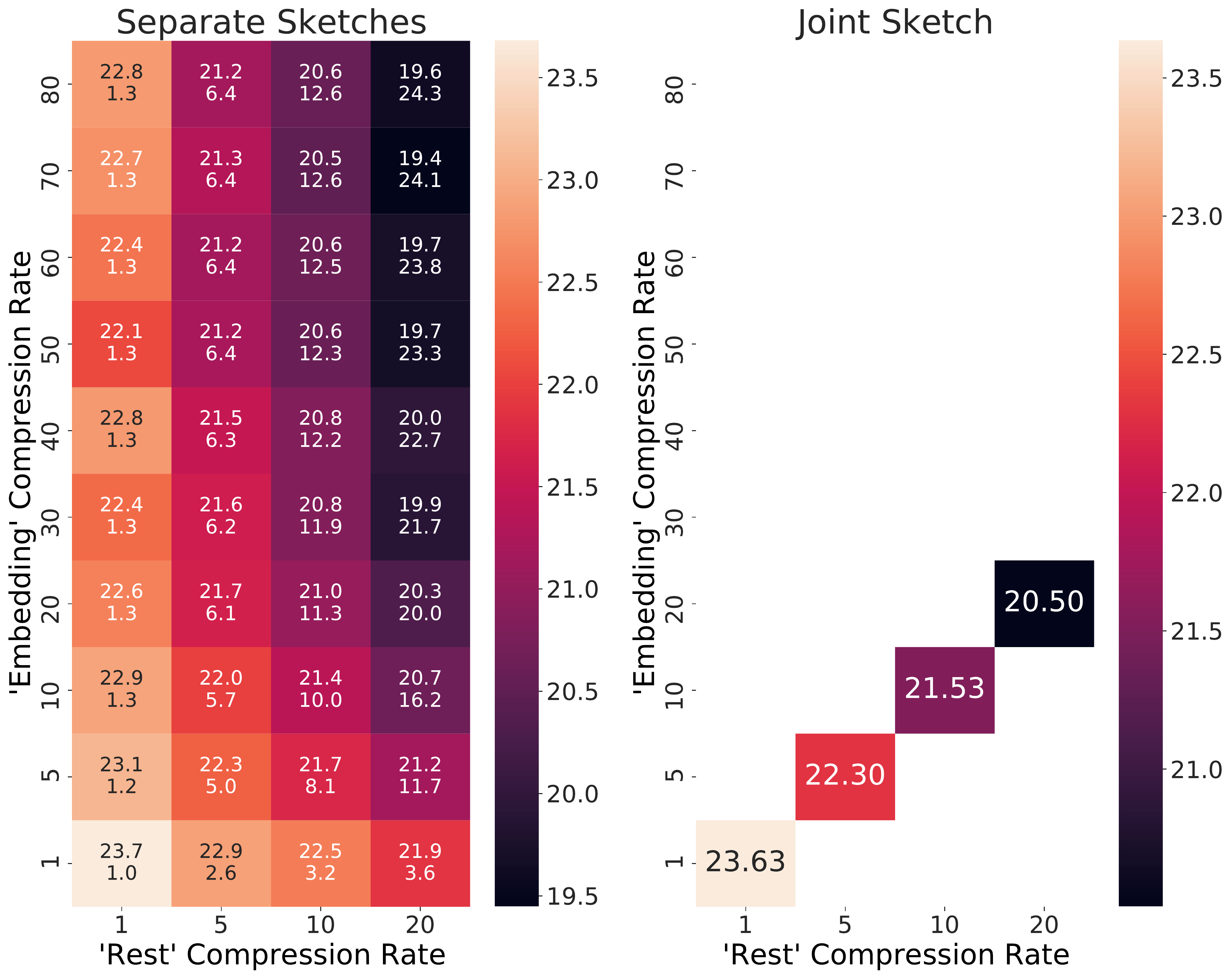}
    \caption{\textbf{Separate sketching does not significantly improve the final model performance.} Heatmap values correspond to the final model test performance followed by (newline) the total compression and are colored by the test performance. Results using the LSTM model on SONWP with $z=0.00$. We train models either by sketching the entire concatenated gradient vector or by sketching the `embedding' layer separate from the `rest' of the model.}
    \label{fig:perlayer-0.0}
\end{figure}

We attempted two additional methods to improve our compression rates. First, we noticed that the LSTM models we trained had consistently different $\ell_2$ norms across layers in training. Because these norms are different, we hypothesizes that sketching and perturbing them separately may improve the model utility. We attempt this protocol in Figures~\ref{fig:perlayer-0.05} and~\ref{fig:perlayer-0.0}, where Figure~\ref{fig:perlayer-0.05} uses $z=0.05$ and Figure\ref{fig:perlayer-0.0} uses $z=0$. We find that, in general, there are no significant performance gains. We further attempt to threshold low values in the sketch. Because this leads to a biased estimate of the gradient, we keep track of the zero-d values in an error term. We tried several threshold values and we found that this as well led to no significant performance gains.

\clearpage
\section{Datasets and Training Setup}\label{app:models}
We run experiments on the full Federated EMNIST and Stack Overflow datasets~\citep{caldas2018leaf}, two common benchmarks for \FL tasks. F-EMNIST has $62$ classes and $N=3400$ clients, with each user holding both a train and test set of examples. In total, there are $671,585$ training examples and $77,483$ test examples. Inputs are single-channel $(28,28)$ images. We sample $n\in[100,1000]$ clients per round for a total $R=1500$ rounds. The Stack Overflow (SO) dataset is a large-scale text dataset based on responses to questions asked on the site Stack Overflow. The are over $10^8$ data samples unevenly distributed across $N=342477$ clients. We focus on the next word prediction (NWP) task: given a sequence of words, predict the next words in the sequence. We sample use $n\in[100,1000]$ and $R=1500$. On F-EMNIST, we experiment with a $\approx 1$ million parameter (4 layer) Convolutional Neural Network  (CNN) used by~\citep{kairouz2021distributed}. On SONWP, we experiment with a $\approx 4$ million parameter (4 layer) long-short term memory (LSTM) model, which is the same as prior work~\cite{andrew2019differentially,kairouz2021distributed}.

On F-EMNIST, we use a server learning rate of $1.$ normalized by $n$ (the number of clients) and momentum of $0.9$~\citep{polyak1964some}; the client uses a learning rate of $0.01$ without momentum. On Stack Overflow, we use a server learning rate of $1.78$ normalized by $n$ and momentum of $0.9$; the client uses a learning rate of $0.3$.

For \DDP, we use the geometric adaptive clipping of~\cite{andrew2019differentially} with an initial $\ell_2$ clipping norm of $0.1$ and a target quantile of $0.5$. We use the same procedure as~\cite{kairouz2021distributed} and flatten using the Discrete Fourier Transform, pick $\beta=\exp{(-0.5)}$ as the conditional randomized rounding bias, and use a modular clipping target probability of $6.33\mathrm{e}{-5}$ or $\approx4$ standard deviations at the server (assuming normally distributed updates). We communicate $16$ bits per parameter for F-EMNIST and $18$ bits for SONWP unless otherwise indicated.

On F-EMNIST, our `large' model corresponds to the CNN whereas our `small' model corresponds to an $\approx 200,000$ parameter model with $3$ dense layers (see Figure~\ref{fig:small-model}). 
\subsection{Model Architectures}
\begin{figure}[!ht]
    \centering
\begin{Verbatim}
Model: "Large Model: 1M parameter CNN"
_________________________________________________________________
 Layer (type)                Output Shape              Param #   
=================================================================
 Conv2D                    (None, 26, 26, 32)          320       
                                                                 
 MaxPooling2D              (None, 13, 13, 32)          0         
                                                                                                            
 Conv2D                    (None, 11, 11, 64)          18496     
                                                                 
 Dropout                   (None, 11, 11, 64)          0         
                                                                 
 Flatten                   (None, 7744)                0         
                                                                 
 Dense                     (None, 128)                 991360    
                                                                 
 Dropout                   (None, 128)                 0         
                                                                 
 Dense                     (None, 62)                  7998      
                                                                 
=================================================================
Total params: 1,018,174
Trainable params: 1,018,174
Non-trainable params: 0
_________________________________________________________________
\end{Verbatim}
    \caption{`Large' model architecture.}
    \label{fig:large-model}
\end{figure}
\begin{figure}[!ht]
    \centering
\begin{Verbatim}
"Small Model: 200k parameter Dense DNN"
_________________________________________________________________
 Layer                Output Shape              Param #   
=================================================================
 Reshape              (None, 784)               0         
                                                                 
 Dense                (None, 200)               157000    
                                                                 
 Dense                (None, 200)               40200     
                                                                 
 Dense                (None, 62)                12462     
                                                                 
=================================================================
Total params: 209,662
Trainable params: 209,662
Non-trainable params: 0
_________________________________________________________________
\end{Verbatim}
    \caption{`Small' model architecture.}
    \label{fig:small-model}
\end{figure}

\begin{figure}[!ht]
    \centering
\begin{Verbatim}
Model: "Stack Overflow Next Word Prediction Model"
_________________________________________________________________
 Layer (type)                Output Shape              Param #   
=================================================================
 InputLayer                (None, None)                0         
                                                                 
 Embedding                 (None, None, 96)            960384    
                                                                 
 LSTM                      (None, None, 670)           2055560   
                                                                 
 Dense                     (None, None, 96)            64416     
                                                                 
 Dense                     (None, None, 10004)         970388    
                                                                 
=================================================================
Total params: 4,050,748
Trainable params: 4,050,748
Non-trainable params: 0
_________________________________________________________________
\end{Verbatim}
    \caption{Stack Overflow Next Word Prediction model architecture.}
    \label{fig:sonwp-model}
\end{figure}
\newpage

\section{Empirical Details of DP and Linear Compression}\label{app:DP}
\paragraph{Noise multiplier to $\mathbf{\varepsilon}$-DP}
We specify the privacy budgets in terms of the noise multiplier $z$, which together with the clients per round $n$, total clients $N$, number of rounds $R$, and the clipping threshold completely specify the trained model $\varepsilon$-DP. Because the final $\mathbf{\varepsilon}$-DP values depend on the sampling method: e.g., Poisson vs. fixed batch sampling, which depends on the production implementation of the FL system, we report the noise multipliers instead. Using~\citet{canonne2020discrete,mironov2017renyi}, our highest noise multipliers roughly correspond to $\varepsilon=\{5,10\}$ using $\delta=1/N$ and privacy amplification via fixed batch sampling.

\paragraph{Linear compression} We display results in terms of the noise multiplier which fully specifies the $\varepsilon$-DP given our other parameters ($n$, $N$, and $R$). We discuss this choice in Appendix~\ref{app:DP}. We use a sparse random projection (i.e. a count sketch together with a mean estimator) which compresses gradients to a sketch matrix of size ($length$,$width$). We test $length\in\{10,15,20,25\}$ and find that $15$ leads to optimal final test performance. We use this value for all our experiments and calculate the $width=d/(r \cdot length)$ where $gradient \in \mathcal{R}^d$ and $r$ is the compression rate. We normalize each sketch row by the $length$ to lower the clipping norm, finding some improvements in our results. Thus, decoding requires only summing the gradient estimate from each row. We provide the full algorithms in below in Appendix~\ref{app:sketching-practice}.

\section{Linear Compression In Practice}\label{app:sketching-practice}
\begin{algorithm}[bh]
  \caption{\textit{Gradient Count-Mean Sketch Encoding.} We find that normalizing (Line 6) in the encoding step improves performance by reducing the norm of the sketch.}
  \label{alg:count-sketch-encode}
  
  \algorithmicrequire{ Gradient vector $g$, sketch width $sw$, sketch length $sl$, shared seed $seed$}

  \begin{algorithmic}[1]
      \STATE $sketch \gets zeros((sl, sw))$
      \FOR{$hash\_index$ in $[0,\cdots,S.length]$, in parallel}
        \STATE $hash\_seed \gets hash\_index + seed$
        \STATE $indices \gets$random\_uniform($0, S.width, hash\_seed$)
        \STATE $signs \gets$ random\_choice($[-1, 1], hash\_seed$)
        \STATE $weights \gets signs \times \frac{grad}{sl}$  %
        \STATE $sketch[hash\_index] \gets $bincount($indices, weights, sl$)
      \ENDFOR
      \STATE \textbf{Return: } $sketch$
  \end{algorithmic}
\end{algorithm}

\begin{algorithm}[bh]
  \caption{\textit{Gradient Count-Mean Sketch Decoding}}
  \label{alg:count-sketch-decode}
  
  \algorithmicrequire{ Sketch $S$, gradient vector size $d$, shared seed $seed$}

  \begin{algorithmic}[1]
      \STATE $gradient\_estimate \gets zeros(d)$
      \FOR{$hash\_index$ in $[0,\cdots,S.length]$, in parallel}
        \STATE $hash\_seed \gets hash\_index + seed$
        \STATE $indices \gets$random\_uniform($0, S.width, hash\_seed$)
        \STATE $signs \gets$ random\_choice($[-1, 1], hash\_seed$)
        \STATE $gradient\_estimate += signs \cdot S[hash\_index, indices]$
      \ENDFOR
      \STATE \textbf{Return: }{$gradient\_estimate$}
  \end{algorithmic}
\end{algorithm}

\clearpage
\section{Additional Details of the Distributed Discrete Gaussian Mechanism}

\begin{algorithm}[tbh]
    \begin{algorithmic}
	\STATE \textbf{Inputs:} Private vector $x_i\in \mbb{R}^d$, Dimension $d$; clipping threshold $c$; granularity $\gamma > 0$;  modulus $M \in \mbb{N}$; noise scale $\sigma > 0$; bias $\beta \in [0, 1)$
	
	\STATE Clip and scale vector: $x_i' = \frac{1}{\gamma}\min(1, \frac{c}{\lV x_i \rV_2})\cdot x_i \in \mbb{R}^d$\;
	\STATE Flatten vector: $x_i'' = H_dDx'_i$ where $H_d$ is the $d$-dim Hadamard matrix and $D$ is a diagonal matrix with each diagonal entry $\msf{unif}\{+1, -1\}$\;
	\STATE Conditional rounding:
	\WHILE{$\lV \tilde{x}_i\rV_2 > \min\Big\{ c/\gamma+\sqrt{d},$ $\sqrt{c^2/\gamma^2+\frac{1}{4}d+\sqrt{2\log(1/\beta)}\lp c/\gamma + \frac{1}{2}\sqrt{d} \rp}\Big\}$}
	\STATE $\tilde{x}_i \in \mbb{Z}^d$ be a randomized rounding of $x_i'' \in \mbb{R}^d$ (i.e. $\E\lb \tilde{x}_i \rb = x_i''$ and $ \lV \tilde{x}_i - x_i'' \rV_\infty \leq 1$)
	\ENDWHILE
	\STATE Perturbation: $z_i = \tilde{x}_i + \mcal{N}_{\mbb{Z}}(0, \sigma^2/\gamma^2) \mod{M}$, where $\mcal{N}_{\mbb{Z}}$ is the discrete Gaussian noise\;
	\STATE\textbf{Return:} $z_i \in \mbb{Z}_M^d$
	\end{algorithmic}
	\caption{Distributed Discrete Gaussian mechanism $\texttt{DDG}_{\msf{enc}}$ (with detailed parameters) \citep{kairouz2021distributed}}\label{alg:ddg_detailed}
\end{algorithm}

\begin{theorem}[private mean estimation with SecAgg \cite{kairouz2021distributed}]\label{thm:ddg}
Define
\begin{align*}
    \Delta^2_2 & \eqDef \min\lbp c^2+\frac{\gamma^2d}{4} + \sqrt{2\log\lp1/\beta\rp}\gamma\lp c+\frac{\gamma}{2}\sqrt{d}\rp, \lp c+\gamma\sqrt{d}\rp^2 \rbp,\\
    \tau & \eqDef 10\sum_{k=1}^{n-1}\exp\lp -2\pi^2\frac{\sigma^2}{\gamma^2}\frac{k}{k+1} \rp,\\
    \varepsilon & \eqDef \min\lbp \sqrt{\frac{\Delta^2_2}{n\sigma^2}+\frac{1}{2}\tau d}, \frac{\Delta_2}{\sqrt{n\sigma}}+\tau\sqrt{d} \rbp,\\
    M & \geq O\lp n+\sqrt{\frac{\varepsilon^2n^3}{d}} + \frac{\sqrt{d}}{\varepsilon} \log\lp n+ \sqrt{\frac{\varepsilon^2n^3}{d}}+ \frac{\sqrt{d}}{\varepsilon}\rp\rp,\\
    \beta &\leq {\Theta}\lp\frac{1}{n} \rp,\\
    \sigma &= \tilde{\Theta}\lp \frac{c}{\varepsilon\sqrt{n}}+ \frac{\gamma\sqrt{d}}{\varepsilon \sqrt{n}}\rp, \\
    \gamma &= \tilde{\Theta}\lp \min\lp \frac{cn}{M\sqrt{d}}, \frac{c}{\varepsilon M} \rp \rp.
\end{align*}
Then Algorithm~\ref{alg:ddg_detailed} satisfies 
\begin{itemize}
    \item $\frac{1}{2}\varepsilon^2$-concentrated differential privacy (which implies $(\alpha, \frac{\varepsilon^2}{\alpha})$-RDP)
    \item $O\lp d \log\lp M\rp \rp = O\lp d\log\lp n+\sqrt{\frac{n^3\varepsilon^2}{d}}+\frac{\sqrt{d}}{\varepsilon}\rp\rp$ bits per-client communication cost;
    \item MSE $\E\lb \lV \hat{\mu} - \mu\rV^2_2 \rb = O\lp \frac{c^2d}{n^2\varepsilon^2} \rp$.
\end{itemize}
\end{theorem}

\section{Proof of Theorem~\ref{thm:random_projection_upper_bound}}\label{proof:random_projection_upper_bound}
Similar as in the proof in Lemma~\ref{lemma:utility_cdp}, let $\mcal{E}$ be the event that $y_i$ is clipped in the $\msf{DDG}$ pre-processing stage for some $i \in [n]$:
$$ \mcal{E} \eqDef \bigcup_{i\in[n]}\lbp \lV Sx_i \rV^2_2\geq 1.1\cdot \lV x_i \rV^2_2 \rbp. $$
By picking $m = \Omega\lp \log\lp \frac{n}{\beta} \rp \rp$ and applying Lemma~\ref{lemma:SJL} together with the union bound, we have $\Pr_S\lbp \mcal{E} \rbp \leq \beta$.

Next, we decompose the error as 
\begin{align*}
\E\lb \lV\hat{\mu} - \mu\rV^2_2 \rb 
&= \E\lb \lV S^\intercal \lp\frac{1}{n}\sum_i\hat{\mu}_y - \frac{1}{n}\sum_i y_i\rp +  \lp S^\intercal\frac{1}{n}\sum_i\hat{\mu}_y - \mu\rp \rV^2_2 \rb\\
& \leq 2\underbrace{\E\lb \lV S^\intercal \lp\hat{\mu}_y - \frac{1}{n}\sum_i y_i\rp\rV^2_2\rb}_{\text{privatization error}}+ 2\underbrace{\E\lb\lV \lp S^\intercal\frac{1}{n}\sum_i y_i - \mu\rp \rV^2_2 \rb}_{\text{compression error}},
\end{align*}
where we use $\hat{\mu}_y$ to denote the output of the $\msf{DDG}$ mechanism. Let us bound the privatization error and the compression error separately. 

\paragraph{Bounding the privatization error} For the first term, observe that $\hat{\mu}_y$ is a function $\lp \msf{clip}\lp y_1 \rp,...,\msf{clip}\lp y_n \rp \rp$, and conditioned on $\mcal{E}^c$, we have 
$$ \hat{\mu}_y\lp \msf{clip}\lp y_1 \rp,...,\msf{clip}\lp y_n \rp \rp = \hat{\mu}_y\lp y_1,...,y_n\rp.$$
For simplicity, let us denote them as $\hat{\mu}_{y, \msf{cl}}$ and $\hat{\mu}_y$ respectively. Next, we separate the error due to clipping by decompose the privatization error into
\begin{align}\label{eq:random_projection_ddg_error}
    \E\lb \lV S^\intercal \lp\hat{\mu}_{y, \msf{cl}} - \frac{1}{n}\sum_i y_i\rp\rV^2_2\rb 
    & \leq \Pr\lbp \mcal{E}^c \rbp \cdot \E\lb \lV S^\intercal \lp\hat{\mu}_y - \frac{1}{n}\sum_i y_i\rp\rV^2_2 \mv \mcal{E}^c\rb + c^2(d+1)\Pr\lbp \mcal{E} \rbp \nonumber\\
    &\leq \E\lb \lV S^\intercal \underbrace{\lp\hat{\mu}_y - \frac{1}{n}\sum_i y_i\rp}_{\text{error due to DDG}}\rV^2_2\rb +c^2(d+1)\beta.
\end{align}

From the MSE bound of DDG \cite{kairouz2021distributed}, we know that with probability $1$,
$$ \E\lb  \lV S^\intercal \lp\hat{\mu}_y - \frac{1}{n}\sum_i y_i\rp \rV^2_2\mv S\rb  = O\lp \frac{c^2m^2}{n^2\varepsilon^2} \rp. $$
Therefore the first term in \eqref{eq:random_projection_ddg_error} can be controlled by Lemma~\ref{lemma:inverse_sketch}, and we can bound the privatization error by 
\begin{align*}
    \E\lb \lV S^\intercal \lp\hat{\mu}_{y,\msf{cl}} - \frac{1}{n}\sum_i y_i\rp\rV^2_2\rb 
    & = O\lp \frac{c^2d}{n^2\varepsilon^2} \rp +c^2(d+1)\beta.
\end{align*}

\paragraph{Bounding the compression error} Next, by Lemma~\ref{lemma:proj_var_bdd}, the compression error can be bounded by
$$ \E\lb\lV \lp S^\intercal\frac{1}{n}\sum_i y_i - \mu\rp \rV^2_2 \rb \leq \frac{2c^2d}{m}. $$

Putting things together, we obtain
\begin{align*}
\E\lb \lV\hat{\mu} - \mu\rV^2_2 \rb 
&\leq C_1 \frac{c^2d}{n^2\varepsilon^2} + \frac{2c^2d}{m}+c^2(d+1)\beta.
\end{align*}
Therefore if we pick $\beta = \frac{1}{n^2\varepsilon^2}$ (so $m$ has to be $\log \lp n^3\varepsilon^2 \rp$), and $ m = n^2 \varepsilon^2$, we have
\begin{equation}
    E\lb \lV \hat{\mu}-\mu \rV^2_2 \rb \leq C_0\frac{c^2 d}{n^2\varepsilon^2}.
\end{equation}

\section{Proof of Theorem~\ref{thm:compression_lb}}
To prove \eqref{eq:compression_lb_general}, we first claim that if there exists a $b$-bit compression scheme $\lp \mcal{C}, \hat{v}\rp$ such that for all $v \in \mb{B}_d(c)$, $ \E\lb \lV \hat{v} - v \rV^2_2 \rb \leq \gamma^2$, then there exists a $\gamma$-covering $C(\gamma)$ of $\mb{B}_d(c) $, such that $\lba C(\gamma) \rba \leq 2^b$. To see this, observe that $\lbp \E\lb \hat{v}\lp \mcal{C}(m)\rp \rb, m \in [2^b] \rbp$ forms a $\gamma$-covering of $\mb{B}_d(c)$. This is because  for any $v \in \mb{B}_d(c)$, it holds that
\begin{align*}
     \lV \E\lb \hat{v}\rb - v \rV^2_2 \overset{\text{(a)}}{\leq}  \E\lb \lV \hat{v} - v \rV^2_2 \rb \leq \gamma^2,
\end{align*}
where (a) holds by Jensen's inequality.

On the other hand, for any $\gamma$-covering of $\mb{B}_d(c)$, we must have $\lba C(c)\rba \geq \frac{\msf{vol}\lp \mb{B}_d(c) \rp}{\msf{vol}\lp \mb{B}_d(\gamma) \rp} = \lp \frac{c}{\gamma} \rp^d$. Thus we conclude that if $2^b \leq \lp \frac{c}{\gamma}\rp^d$, then $ \E\lb \lV \hat{\mu} - \mu \rV^2_2 \rb \geq \gamma^2$, or equivalently 
$$ \E\lb \lV \hat{\mu} - \mu \rV^2_2 \rb \geq \lp  \frac{1}{2^b}\rp^{2/d}c^2. $$

To prove \eqref{eq:compression_lb_unbiased}, we first impose a product Bernoulli distribution on $\mb{B}_d(c)$, upper bound the \emph{quantized} Fisher information \cite{barnes2019lower}, and then apply the Cramer-Rao lower bound.

To begin with, let $X \sim \prod_{i\in[d]} \msf{Ber}\lp \theta_i \rp$ for some $\theta_i \in [0, 1]$. Then $\frac{c}{\sqrt{d}} X \subset \mb{B}_d(c)$ almost surely. Next, we claim that for any $b$-bit unbiased compression scheme $\lp \mcal{C}, \hat{v}\rp$ such that$ \E\lb \lV \hat{v} - v \rV^2_2 \rb \leq \gamma^2$ for all $v \in \mb{B}_d(c)$,  $\hat{\theta}\lp X \rp \eqDef \frac{\sqrt{d}}{c}\hat{v}\lp \mcal{C}\lp \frac{c}{\sqrt{d}} X \rp \rp$ is an unbiased estimator of $\theta = \lp \theta_1,...,\theta_d \rp \in [0, 1]^d$ with estimation error bounded by
$$ \max_{\theta \in [0, 1]^d}\E\lb \lV \hat{\theta}\lp X \rp - \theta\rV^2_2 \rb \leq \frac{d\gamma}{c}. $$
To see this, observe that
\begin{align}\label{eq:crlb_upbdd}
    \E\lb \lV \frac{\sqrt{d}}{c}\hat{v}\lp \mcal{C}\lp \frac{c}{\sqrt{d}} X \rp \rp - \theta\rV^2_2 \rb 
    & \leq 2 \E\lb\lV  \frac{\sqrt{d}}{c}\hat{v}\lp \mcal{C}\lp \frac{c}{\sqrt{d}} X \rp \rp -  X \rV^2_2\rb + 2\E\lb\lV X- \theta\rV^2_2 \rb\nonumber\\
    & \overset{\text{(a)}}{\leq} 2\frac{d\gamma^2}{c^2} + 2\E\lb \lV X - \theta \rV^2_2\rb \nonumber\\
    & \overset{\text{(b)}}{\leq} 2\frac{d\gamma^2}{c^2} + 2\sum_{i\in[d]}\theta_i(1-\theta_i)\nonumber \\
    & \overset{\text{(c)}}{\leq}  2d\lp\frac{\gamma^2}{c^2}+1 \rp,
\end{align}
where (a) holds since by assumption $\E\lb \lV \hat{v} - v \rV^2_2 \rb \leq \gamma $, (b) holds since $X_i \sim \msf{Ber}\lp \theta_i \rp$, and (c) holds since $\theta_i \in [0, 1]$.

Next, we apply \citep[Corollary~4]{barnes2019lower}, which states that for any b bits (possibly randomized) transform $M:\{0,1\}^d\ra \mcal{Y}$ with $\lba \mcal{Y} \rba \leq 2^b$, the Fisher information $I_Y(\theta)$ with $Y\sim M(\cdot|X)$ and $X\sim \prod_{i\in[d]}\msf{Ber}(\theta_i)$ is upper bounded by
$$ \min_{M(\cdot|X)}\max_{\theta \in [0, 1]^d} \msf{Tr}\lp I_Y(\theta)\rp \leq  C_1 \min(d, b).$$

Therefore, since $\mcal{C}$ is a $b$-bit compression operator (and thus $\hat{\theta}$ can be encoded into $b$ bits too), we must have 
$$ \max_{\theta \in [0, 1]^d} \msf{Tr}\lp I_{\hat{\theta}}(\theta)\rp \leq  C_1 \min(d, b).$$
Applying Cramer-Rao lower bound yields
\begin{equation}\label{eq:crlb}
    \max_{\theta \in [0, 1]^d} \E\lb \lV \hat{\theta} - \theta \rV^2_2\rb\geq \sum_{i\in[d]} \lb I_{\theta}(\theta)\rb_{i,i} \geq \frac{d^2}{\msf{Tr}\lp I_{\hat{\theta}}\lp \theta \rp \rp} \geq C_2 \frac{d^2}{\min(b, d)}. 
\end{equation}

Finally, by \eqref{eq:crlb_upbdd} and \eqref{eq:crlb}, we must have
\begin{align*}
     2d\lp\frac{\gamma^2}{c^2}+1 \rp \geq C_2\frac{d^2}{\min\lp b, d \rp} \Longleftrightarrow \gamma^2 \geq \lp C_2\frac{d}{\min\lp b, d \rp} -2\rp
     \geq C_3\frac{dc^2}{\min(b, d)},
\end{align*}
for some constants $C_2, C_3 >0$ and $b \leq \frac{d}{2C_2}$, completing the proof.

\section{Proof of Theorem~\ref{thm:compressed_sensing_ddp}}

\begin{figure}[ht]
\begin{center}
{\includegraphics[width=0.69\linewidth]{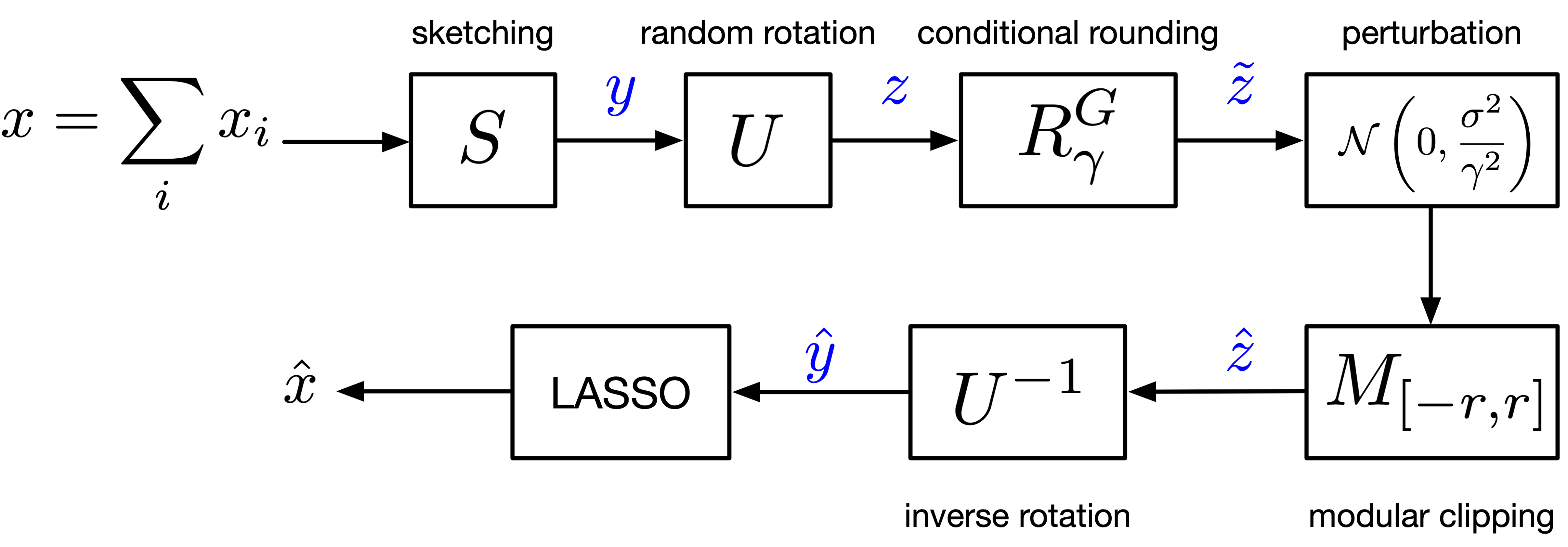}}
\caption{Sparse private aggregation.}
\label{fig:ddg_steps}
\end{center}
\end{figure}

The $\frac{1}{2}\varepsilon^2$-concentrated DP is guaranteed by the DDG mechanism. Therefore we only need to analyze the $\ell_2$ error $\E\lb \lV \hat{\mu} - \mu \rV^2_2 \rb$. To analyze the $\ell_2$ error, we can equivalently formulate it as a sparse linear problem:
$$ \hat{\mu}_y = S\mu + \Delta, $$
where $\Delta = \hat{\mu}_y - \frac{1}{n} \sum_i Sx_i$ is the error introduced by the DDG mechanism. We illustrate each steps of the end-to-end transform of Algorithm~\ref{alg:sparse_DME_lasso} in Figure~\ref{fig:ddg_steps}.

Before we continue to analyze the error, we first introduce some necessary definitions.

\begin{definition}[Restricted eigenvalue (RE) condition \citep{raskutti2010restricted}]
A matrix $S \in \mbb{R}^{m\times d}$ satisfies the restricted eigenvalue (RE) condition over $\mcal{S} \subseteq [d]$ with parameter $(\kappa, \alpha)$ if 
$$ \frac{1}{m}\lV S\Delta \rV^2_2 \geq \kappa \lV \Delta \rV^2_2,  \, \forall \Delta \in \msf{C}_\alpha(\mcal{S}),$$
where $\msf{C}_{\alpha}\lp \mcal{S} \rp \eqDef \lbp \Delta\in\mbb{R}^d | \lV \Delta_{\mcal{S}^c} \rV_1 \leq \alpha\lV \Delta_\mcal{S} \rV_1 \rbp$. If the RE condition holds uniformly for all subsets $\mcal{S}$ with cardinality $s$, we say $S$ satisfies a RE condition of order $s$ with parameters $(\alpha, \kappa)$.
\end{definition}

\begin{definition}[A sufficient condition of RE (a ``soft'' RE)]\label{def:soft_re}
We say $S$ satisfies a ``soft'' RE with parameter $(\kappa, \rho)$, if
\begin{equation}\label{eq:soft_re}
    \frac{1}{m}\lV S \Delta \rV^2_2 \geq \frac{1}{8}\kappa \lV \Delta \rV^2_2 - 50 \rho^2 \frac{\log (d)}{m} \lV \Delta \rV^2_1,\, \text{ for all } \Delta \in \mbb{R}^d.
\end{equation}
\end{definition}

\begin{remark}
Let $S_{i,j}\diid \mcal{N}(0,1)$. Then $S$ satisfies \eqref{eq:soft_re} with $\kappa = \rho = 1$ with probability at least $1-\frac{e^{-m/32}}{1-e^{-m/32}}$. %
\end{remark}

\begin{theorem}[Lasso oracle inequality ( Theorem~7.19 \citet{wainwright2019high})]\label{thm:lasso_oracle}
As long as $S$ satisfies \eqref{eq:soft_re} with $(\kappa, \rho)$ and $\lambda_n \geq \lV S^\intercal\lp \hat{\mu}_y - \mu_y \rp \rV_\infty/m$ then following MSE bound holds:
$$ \lV \hat{\mu} - \mu \rV^2_2 \leq  \frac{144|S|}{c_1^2\kappa^2}\lambda_n^2 +\frac{16}{c_1\kappa}\lambda_n\lV \mu_{S^c} \rV_1 + \frac{32c_2\rho^2}{c_1\kappa}\frac{\log d}{m}\lV \mu_{S^c} \rV^2_1, $$
for all $S \subseteq [d]$ with $\lba S \rba \leq \frac{c_1}{64c_2}\frac{m}{\log d}$. For the Gaussian ensembles, we have $\kappa = \rho = 1$.
\end{theorem}

Therefore according to Theorem~\ref{thm:lasso_oracle}, to upper bound the estimation error, it suffices to control $\frac{\lV S^\intercal\lp \hat{\mu}_y - \mu_y \rp \rV_\infty}{m}$ (and hence the regularizer $\lambda_n$). In the next lemma, we give an upper bound on it.
\begin{lemma}\label{lemma:cs}
Let $d' = c_0 s \log d$, $c' = c\sigma_\msf{max}(S)$ and let granularity $\gamma > 0$, modulus  $M \in \mbb{N}$, noise scale $\sigma > 0$, and bias $\beta \in [0, 1)$ be defined as in Theorem~1 and Theorem~2 in \cite{kairouz2021distributed}. Then as long as
$$ M \geq \frac{2}{\gamma} \sqrt{\lp n\lp \gamma^2+4\sigma^2\rp + \frac{4n^2c'^2}{m}\rp\lp \log m + \log\lp \frac{1}{(1-\beta)^n} \rp +\log\lp\frac{8}{\delta}\rp\rp} ,$$
the following bound holds with probability at least $1-\delta$:
\begin{align}\label{eq:bdd1}
    \lV\frac{S^\intercal \lp \hat{\mu}_y - \mu_y  \rp}{m}\rV_\infty \leq &\sqrt{\frac{1}{n}\lp \log\lp \frac{d}{(1-\beta)^n}+\log\lp \frac{2}{\delta} \rp \rp \lp \frac{\gamma^2+4\sigma^2}{8} \rp \lp \frac{\max_{i=1,...,d}\lV S_i\rV^2_2}{m} \rp \rp}
\end{align}
\end{lemma}

\begin{corollary}
Let each row of $S$ be generated according to $\mcal{N}\lp 0, \mbb{I}_d \rp$ and let $\gamma, \sigma, \beta$ be the parameters used in the discrete Gaussian mechanism. If $|x|_0\leq s$ and
\begin{align}%
    \lambda_n = \sqrt{\frac{1}{n}\lp \log\lp \frac{d}{(1-\beta)^n}+\log\lp \frac{2}{\delta} \rp \rp \lp \frac{\gamma^2+4\sigma^2}{8} \rp \lp \frac{\max_{i=1,...,d}\lV S_i\rV^2_2}{m} \rp \rp},
\end{align}
then with probability at least $1-\delta$, 
$$ \lV \hat{\mu} - \mu \rV^2_2 = O\lp\frac{s}{n}\lp \log\lp \frac{d}{(1-\beta)^n}+\log\lp \frac{2}{\delta} \rp \rp \lp \frac{\gamma^2+4\sigma^2}{8} \rp \lp \frac{\max_{i=1,...,d}\lV S_i\rV^2_2}{m} \rp \rp \rp.$$
In addition, the communication cost is
$$ O\lp  s\log d \log\lp\frac{1}{\gamma} \sqrt{\lp n\lp \gamma^2+4\sigma^2\rp + \frac{4n^2c'^2}{m}\rp\lp \log m + \log\lp \frac{1}{(1-\beta)^n} \rp +\log\lp\frac{8}{\delta}\rp\rp}\rp\rp $$
\end{corollary}

\subsection{Parameter selection}
Now we pick parameters so that Algorithm~1 satisfies $\frac{1}{2}\varepsilon^2$-concentrated DP and attains the MSE as in the centralized model. To begin with, we first determine $\sigma$ so that Algorithm~1 satisfies differential privacy. 
\paragraph{Privacy analysis} To analyze the privacy guarantees, we treat the inner discrete Gaussian mechanism as a black box with the following parameters: effective dimension $d' = m$, $\ell_2$ bound (or the clipping threshold) $c' = c\sigma_{\msf{max}}(S)$, granularity $\gamma > 0$, noise scale $\sigma >0$, and bias $\beta \in [0, 1)$. Define 
\begin{align*}
    \Delta^2_2 & \eqDef \min\lbp c'^2+\frac{\gamma^2d'}{4} + \sqrt{2\log\lp1/\beta\rp}\gamma\lp c'+\frac{\gamma}{2}\sqrt{d'}\rp, \lp c'+\gamma\sqrt{d'}\rp^2 \rbp\\
    \tau & \eqDef 10\sum_{k=1}^{n-1}\exp\lp -2\pi^2\frac{\sigma^2}{\gamma^2}\frac{k}{k+1} \rp\\
    \varepsilon & \eqDef \min\lbp \sqrt{\frac{\Delta^2_2}{n\sigma^2}+\frac{1}{2}\tau d'}, \frac{\Delta_2}{\sqrt{n\sigma}}+\tau\sqrt{d'} \rbp.
\end{align*}
Then Theorem~1 in \cite{kairouz2021distributed} ensures that Algorithm~1 is $\frac{1}{2}\varepsilon^2$-concentrated DP. With this theorem in hands, we first determine $\sigma$. Observe that
$$ \varepsilon^2 \leq \frac{\Delta^2_2}{n\sigma^2}+\frac{1}{2}\tau d' \leq \frac{2c'^2}{n\sigma^2} + \frac{2d'}{n(\sigma/\gamma)^2} + 5nd'\exp^{-\pi^2(\sigma/\gamma)^2}. $$
Thus it suffices to set $\sigma = \max\lbp \frac{2c'}{\varepsilon\sqrt{n}}, \frac{\gamma\sqrt{8d'}}{\varepsilon \sqrt{n}}, \frac{\gamma}{\pi^2}\log\lp \frac{20nd'}{\varepsilon^2} \rp \rbp = \tilde{\Theta}\lp \frac{c'}{\varepsilon\sqrt{n}}+\sqrt{\frac{d'}{n}}\frac{\gamma}{\varepsilon} \rp$.

\paragraph{Accuracy analysis} We set $\beta = \min\lp \sqrt{\frac{\gamma}{n}}, \frac{1}{n} \rp$, and together with the upper bound \eqref{eq:bdd1} and 
$$ \sigma^2 \preceq \frac{c'^2+\gamma^2d'}{n\varepsilon^2}+\gamma^2\log^2\lp \frac{nd'}{\varepsilon^2} \rp, $$
we have
$$ \lambda_n = \Theta\lp \sqrt{\frac{1}{n}\lp \log d + \log(1/\delta) \rp\lp \frac{c'^2+\gamma^2d'}{n\varepsilon^2}+\gamma^2\log^2\lp \frac{nd'}{\varepsilon^2} \rp + \gamma^2\rp \lp  \frac{\max_{i=1,...,d}\lV S_i\rV^2_2}{m}  \rp}\rp. $$

Thus it suffices to pick
$$ \gamma^2 = O\lp \min\lp \frac{c'^2}{d'}, \frac{c'^2}{n\varepsilon^2\log^2\lp \frac{nd'}{\varepsilon^2} \rp} \rp\rp. $$

We summarize the above parameter selection in the following theorem.
\begin{theorem}
By selecting 
\begin{align*}
    \beta &= \min\lp \sqrt{\frac{\gamma}{n}}, \frac{1}{n} \rp\\
    \sigma &= \max\lbp \frac{2c'}{\varepsilon\sqrt{n}}, \frac{\gamma\sqrt{8d'}}{\varepsilon \sqrt{n}}, \frac{\gamma}{\pi^2}\log\lp \frac{20nd'}{\varepsilon^2} \rp \rbp \\
    \gamma^2 &= O\lp \min\lp \frac{c'^2}{d'}, \frac{c'^2}{n\varepsilon^2\log^2\lp \frac{nd'}{\varepsilon^2} \rp} \rp\rp\\
    \lambda_n &= \Theta\lp \sqrt{\frac{1}{n}\lp \log d + \log(1/\delta) \rp\lp \frac{c'^2}{n\varepsilon^2}\rp \lp  \frac{\max_{i=1,...,d}\lV S_i\rV^2_2}{m}  \rp}\rp = O\lp \frac{c\log d}{n\varepsilon} \rp\\
    M &= O\lp \sqrt{\lp \frac{d'}{\varepsilon^2}+ n^2 \rp\lp \log d' + \log\lp1/\delta\rp \rp} \rp 
\end{align*}
Algorithm~\ref{alg:ddg_detailed} is $\frac{1}{2}\varepsilon^2$-concentrated DP, and with probability at least $1-\delta$,
\begin{align*}
    \lV \hat{\mu} - \mu \rV^2_2 
    & = O\lp\frac{s(\log d+\log(1/\delta))c'^2}{n^2\varepsilon^2}\lp \frac{\max_{i\in[d]}\lV S_i\rV^2_2}{d'} \rp \rp \\
    & = O\lp\frac{s(\log d+\log(1/\delta))c^2\rho^2_{\msf{max}}(S)}{n^2\varepsilon^2}\lp \frac{\max_{i\in[d]}\lV S_i\rV^2_2}{s\log(d)} \rp \rp.
\end{align*}
\end{theorem}

This establishes Theorem~\ref{thm:compressed_sensing_ddp}.

\section{Proof of Lemmas}
\subsection{Proof of Lemma~\ref{lemma:inverse_sketch}}\label{sec:proof_lemma_inverse_sketch}

First, we break $v \in \mbb{R}^m$ into $t$ blocks
$$ v = \begin{bmatrix}
v^{(1)}\\
v^{(2)}\\
\vdots\\
v^{(t)}
\end{bmatrix},$$
where $v_j \in \mbb{R}^w$ (recall that $m = w\cdot t$). Then
\begin{align*}
    \E\lb\lV S^\intercal v \rV^2_2\rb
    & = \frac{1}{t}\E\lb\lV \sum_{i=1}^t S_i^\intercal v^{(i)} \rV^2_2\rb\\
    & = \frac{1}{t}\E\lb\lV \sum_{i=1}^t \lp S_i^\intercal v^{(i)} - \E\lb S_i^\intercal v^{(i)}\rb \rp + \sum_{i=1}^t\E\lb S_i^\intercal v^{(i)} \rb \rV^2_2\rb\\
    & \leq \frac{2}{t}\lp \sum_{i=1}^t\E\lb\lV S_i^\intercal v^{(i)} -\E\lb S_i^\intercal v^{(i)} \rb\rV^2_2\rb+\lV \sum_{i=1}^t\E\lb S_i^\intercal v^{(i)} \rb \rV^2_2 \rp \\
    & \leq \underbrace{\frac{2}{t} \sum_{i=1}^t\E\lb\lV S_i^\intercal v^{(i)} \rV^2_2\rb}_{\text{(a)}}+\underbrace{\frac{2}{t}\lV \sum_{i=1}^t\E\lb S_i^\intercal v^{(i)} \rb \rV^2_2 }_{\text{(b)}}.
\end{align*}
Now we bound each term separately. To bound (a), observe that for all $i \in [t]$,
$$ \E\lb\lV S_i^\intercal v^{(i)} \rV^2_2\rb \leq \E\lb N(S_i) \E\lb\lV v^{(i)}\rV^2_2\mv S_i\rb \rb\leq \E\lb N(S_i) \rb B^2, $$
since by assumption $\E\lb\lV v^{(i)}\rV^2_2\mv S_i\rb \leq B^2$ almost surely, where $N(S_i)$ is the maximum amount of 1s in $w$ rows of $S_i$. Notice that this amount is the same as the maximum load of throwing $d$ balls into $w$ bins. Applying a Chernoff bound, this quantity can be upper bounded by 
$ \E[N(S_i)] \leq \frac{(e+1)d}{w}, $
so (a) is bounded by
\begin{align*}
    \frac{2}{t} \sum_{i=1}^t\E\lb\lV S_i^\intercal v^{(i)} \rV^2_2\rb 
    & \leq \frac{2(e+1)d}{w t}\sum_{i=1}^t\lV v^{(i)} \rV^2_2 \leq \frac{8d}{m}\lV v \rV^2_2.
\end{align*}

To bound (b), observe that
\begin{align*}
    \E\lb S_i^\intercal v^{(i)} \rb %
    = \frac{1}{w} \begin{bmatrix}
    \sum_{j=1}^w v^{(i)}_j \\
    \sum_{j=1}^w v^{(i)}_j\\
    \vdots \\
    \sum_{j=1}^w v^{(i)}_j
    \end{bmatrix} = \lp \frac{1}{w}\sum_{j=1}^w v^{(i)}_j\rp\cdot \mb{1}_d,
\end{align*}
where $\mb{1}_d \eqDef [1, ..., 1]^\intercal \in \mbb{R}^d$. Therefore, summing over $i\in[t]$, we have
\begin{align*}
    \sum_{i=1}^t\E\lb S_i^\intercal v^{(i)} \rb =  \sum_{i=1}^t\lp \frac{1}{w}\sum_{j=1}^w v^{(i)}_j\rp\cdot \mb{1}_d
    = \frac{1}{w} \lp \sum_{i=1}^m v_i \rp \cdot\mb{1}_d.
\end{align*}
Thus we can bound (b) by
\begin{align*}
    \frac{2}{t}\lV \sum_{i=1}^t\E\lb S_i^\intercal v^{(i)} \rb \rV^2_2 
    \leq \frac{2d}{tw}\lp \frac{\lp \sum_{i=1}^m v_i \rp^2}{w} \rp \leq \frac{2d}{m}\lV v \rV^2_2,
\end{align*}
where the last inequality follows from the Cauchy-Schwartz inequality.

Putting (a) and (b) together, the proof is complete.

\subsection{Proof of Lemma~\ref{lemma:utility_cdp}}\label{proof:utility_cdp}
For simplicity, let $\mu \eqDef \frac{1}{n}\sum_i x_i$, and $N \sim \mcal{N}\lp 0, \sigma^2I_m \rp$. Define 
$$ \mcal{E}_\alpha \eqDef \bigcup_{i\in[n]}\lbp \lV Sx_i \rV^2_2\geq (1+\alpha)\cdot \lV x_i \rV^2_2 \rbp. $$
We will pick $m = \Omega\lp \frac{1}{\alpha^2}\log\lp \frac{n}{\beta} \rp \rp$, so by Lemma~\ref{lemma:SJL} and the union bound $\Pr_S\lbp \mcal{E}^c_\alpha \rbp \leq \beta$.
Then the MSE can be computed as
\begin{align*}
    &\E\lb \lV S^\intercal \lp\frac{1}{n}\sum_i\msf{clip}\lp S x_i \rp+N\rp - \mu \rV^2_2 \rb\\
    & \overset{\text{(a)}}{=} \E\lb \lV S^\intercal \lp\frac{1}{n}\sum_i\msf{clip}\lp S x_i \rp\rp - \mu \rV^2_2 \rb + \E\lb \lV S^\intercal N \rV^2_2 \rb \\
    & \overset{\text{(b)}}{\leq} \E\lb \lV S^\intercal \lp\frac{1}{n}\sum_i\msf{clip}\lp S x_i \rp\rp - \mu \rV^2_2 \mv \mcal{E}_\alpha^c\rb\cdot \Pr\lbp \mcal{E}_\alpha^c \rbp + (d+1)\cdot\Pr\lbp \mcal{E}_\alpha \rbp+\E\lb \lV S^\intercal N \rV^2_2 \rb\\
    & \overset{\text{(c)}}{\leq} \E\lb \lV \frac{1}{n}\sum_i S^\intercal S x_i - \mu \rV^2_2\rb + (d+1)\beta+\E\lb \lV S^\intercal N \rV^2_2 \rb,
\end{align*}
where (a) holds since $\E\lb S^\intercal N \mv S \rb = 0$ almost surely, (b) holds since  $ \lV S^\intercal \nu\rV^2_2 \leq d$ for all count-sketch matrix $S$ and all $\lV \nu\rV_2 \leq 1$ (so $ \lV S^\intercal \lp\frac{1}{n}\sum_i\msf{clip}\lp Sx_i \rp \rp - \mu \rV^2_2 \leq d+1 $), and (c) holds since conditioned on $\mcal{E}_\alpha^c$, $\msf{clip}\lp Sx_i \rp = Sx_i$ for all $i$.

Next, we control each term separately. The first term can be controlled using Lemma~\ref{lemma:proj_var_bdd}, which gives
\begin{equation*}
    \E\lb \lV \frac{1}{n}\sum_i S^\intercal S x_i - \mu \rV^2_2\rb \leq\frac{2d}{m}.
\end{equation*}

The third term can be computed as follows:
\begin{align*}
    \E\lb \lV S^\intercal\cdot N \rV^2_2\rb = \E\lb \E\lb S^\intercal\cdot N \mv S\rb \rb \leq \sigma^2 d = \frac{8d(1+\alpha)\log\lp 1.25 \rp}{n^2\varepsilon^2}.
\end{align*}
Thus we arrive at
$$ \E\lb \lV \hat{\mu}-\mu \rV^2_2 \rb \leq \frac{2d}{m} + (d+1)\beta + \frac{8d(1+\alpha)\log(1.25)}{n^2\varepsilon^2}. $$

Therefore, if we pick $\beta = \frac{\kappa}{n^2\varepsilon^2}$ (so $m = \Omega\lp \frac{1}{\alpha^2}\log\lp \frac{n^3\varepsilon^2}{\kappa} \rp \rp$), and $ m = \frac{n^2\varepsilon^2}{\kappa}$, we have
\begin{equation}
    E\lb \lV \hat{\mu}-\mu \rV^2_2 \rb \leq \frac{8d\log(1.25)+ d\lp 8\alpha\cdot \log(1.25)+ \lp 3+\frac{1}{d}\rp\kappa\rp}{n^2\varepsilon^2}.
\end{equation}

Notice that we can make $\alpha$ and $\kappa$ small, say $\alpha = \kappa = 0.1$, and the MSE will be closed to the uncompressed one.

\subsection{Proof of Lemma~\ref{lemma:cs}}

We first set some notation. Let $\mu_y =S\mu = \frac{1}{n}\sum_i y_i$, $z = U\mu_y$ (where $U$ is the random rotation matrix), $\tilde{z} = \frac{1}{n}\sum_i\lp R^G_\gamma(z_i) + \mcal{N}_{\mbb{Z}}(0, \sigma^2/\gamma^2)\rp$, $\hat{z} = \frac{1}{n}M_{[-r,r]}(n\tilde{z})$ and finally $\hat{\mu}_y = U^\intercal \hat{z}$,
where $R^G_\gamma(\cdot)$ is the randomized rounding, $\mcal{N}_{\mbb{Z}}$ is the discrete Gaussian noise, and $M_{[-r,r]}(\cdot)$ is the module clipping (details can be found in \cite{kairouz2021distributed}).

Now, we can write the left-hand side of \eqref{eq:bdd1} as 
$$  \lV\frac{S^\intercal \lp \hat{\mu}_y - \frac{1}{n}\sum_i y_i \rp}{m}\rV_\infty = \lV\frac{S^\intercal U^\intercal \lp \hat{z} - \tilde{z} + \tilde{z} - z \rp}{m}\rV_\infty \leq  \underbrace{\lV\frac{S^\intercal U^\intercal \lp \hat{z} - \tilde{z}\rp}{m}\rV_\infty}_{\text{(1): module error}} + \underbrace{\lV\frac{S^\intercal U^\intercal \lp \tilde{z} -z \rp}{m}\rV_\infty}_{\text{(2): rounding error}}.$$

Define $S' \eqDef \frac{S^\intercal U^\intercal}{\sqrt{m}}$ and let $S'_i$ be the $i$-th row of $S'$ for all $i=1,...,d$. Now we bound (1) and (2) separately.

\paragraph{Bounding the module error}
Observe that since $\lV y \rV^2_2 \leq c'^2$, by Lemma~30 in \cite{kairouz2021distributed} we have
$$ \forall t\in\mbb{R} \forall j \in [m] \, \E\lb \exp\lp tz_j \rp \rb =  \E\lb \exp\lp t\lp U n\mu_y\rp_j \rp \rb \leq \exp\lp\frac{t^2n^2c'^2}{2m}\rp.$$

Applying the union bound and the Chernoff's bound yields 
\begin{equation}\label{eq:z1}
    \Pr\lbp \max_{j=1,...,m} \lba z_j \rba > \sqrt{2\frac{n^2c'^2}{m}\lp \log m + \log\lp \frac{8}{\delta} \rp \rp}\rbp \leq \frac{\delta}{4},
\end{equation}
where the randomness is over  the random rotational matrix $U$.

On the other hand, from Proposition~26, we have
$$ \E\lb \exp\lp t\lp n\tilde{z}_j - nz_j\rp \rp \rb \leq \frac{\exp\lp \frac{nt^2\lp \gamma^2 + 4 \sigma^2\rp}{8} \rp}{\lp 1- \beta \rp^n}.  $$
Applying the Markov's inequality and the union bound, we obtain
$$  \Pr\lbp\max_{j\in[m]} \lba n\tilde{z}_j - nz_j\rba \geq t \rbp \leq \frac{m}{(1-\beta)^n}\exp\lp -\frac{2t^2}{n\gamma^2 + 4\delta^2} \rp. $$
Thus picking $t = \sqrt{\frac{n\lp\gamma^2+4\sigma^2\rp}{2}\lp \log\lp \frac{1}{(1-\beta)^n} \rp +\log\lp\frac{4}{\delta}\rp\rp} $ yields
\begin{equation}\label{eq:z2}
    \Pr\lbp\max_{j\in[m]} \lba n\tilde{z}_j - nz_j\rba \geq \sqrt{\frac{n\lp\gamma^2+4\sigma^2\rp}{2}\lp \log\lp \frac{1}{(1-\beta)^n} \rp +\log\lp\frac{4}{\delta}\rp\rp} \rbp \leq \frac{\delta}{4}. 
\end{equation}

Putting \eqref{eq:z1} and \eqref{eq:z2} together, we arrive at
\begin{equation}\label{eq:z3}
    \Pr\lbp\max_{j\in[m]} \lba \tilde{z}_j \rba \geq \sqrt{\lp n\lp\gamma^2+4\sigma^2\rp + \frac{4n^2c'^2}{m}\rp\lp \log m + \log\lp \frac{1}{(1-\beta)^n} \rp +\log\lp\frac{8}{\delta}\rp\rp} \rbp \leq \frac{\delta}{2}, 
\end{equation}
where we use the fact that $\sqrt{a} + \sqrt{b} \leq \sqrt{2(a+b)}$ and the union bound.

Finally, observe that as long as $\lV \tilde{z} \rV_\infty \leq r$, $\hat{z} \eqDef M_{[-r, r]}(\tilde{z}) = \tilde{z}$. Thus by picking 
$$ r\eqDef \sqrt{\lp n\lp \gamma^2+4\sigma^2\rp + \frac{4n^2c'^2}{m}\rp\lp \log m + \log\lp \frac{1}{(1-\beta)^n} \rp +\log\lp\frac{8}{\delta}\rp\rp}, $$
we have 
$$ \Pr\lbp \lV\frac{S^\intercal U^\intercal \lp \hat{z} - \tilde{z}\rp}{m}\rV_\infty > 0 \rbp  \leq \frac{\delta}{2}. $$

\paragraph{Bounding the module error}

First notice that 
\begin{align*}
    \Pr\lbp \lV\frac{S^\intercal U^\intercal \lp n\hat{z} - n\tilde{z}\rp}{m}\rV_\infty \geq t \rbp 
    & = \Pr\lbp \max_{i\in[d]} \lan S'_i, (n\tilde{z} - nz) \ran \geq t \rbp + \Pr\lbp \max_{i\in[d]} -\lan S'_i, (n\tilde{z} - nz) \ran \geq t \rbp 
\end{align*}
Thus we have
\begin{align*}
   \Pr\lbp \max_{i\in[d]} \lan S'_i, (n\tilde{z} - nz) \ran \geq t \rbp  
   & \leq \sum_{i\in[d]} \Pr\lbp \exp\lp \lan \lambda S'_i, nz-n\tilde{z}\ran\rp \geq \exp\lp \lambda t \rp \rbp \\
   & \overset{\text{(a)}}{\leq} \sum_{i\in[d]}\frac{\exp\lp \frac{n(\gamma^2+4\sigma^2)}{8}\lV S_i' \rV^2_2 \lambda^2 - \lambda t \rp}{(1-\beta)^n} \\
   & \overset{\text{(b)}}{\leq} \frac{d}{(1-\beta)^n} \exp\lp -\frac{2mt^2}{n\lp \gamma^2 + 4\sigma^2 \rp \max_{i \in [d]}\lV S_i\rV^2_2} \rp,
\end{align*}
where (a) holds by Proposition~26 in \cite{kairouz2021distributed}, and (b) holds by picking $\lambda$ properly. 

On the other hand, 
\begin{align*}
    \Pr\lbp \max_{i\in[d]} -\lan S'_i, (n\tilde{z} - nz) \ran \geq t \rbp 
    & \leq \sum_{i\in[d]} \Pr\lbp \exp\lp \lan -\lambda S'_i, nz-n\tilde{z}\ran\rp \geq \exp\lp \lambda t \rp \rbp \\
    & \overset{\text{(a)}}{\leq}\frac{d}{(1-\beta)^n} \exp\lp -\frac{2mt^2}{n\lp \gamma^2 + 4\sigma^2 \rp \max_{i \in [d]}\lV S_i\rV^2_2} \rp,
\end{align*}
where (a) holds due to the same reason.

Thus picking
$$ t = \sqrt{n\lp \log\lp \frac{d}{(1-\beta)^n}+\log\lp \frac{4}{\delta} \rp \rp \lp \frac{\gamma^2+4\sigma^2}{8} \rp \lp \frac{\max_{i=1,...,d}\lV S_i\rV^2_2}{m} \rp \rp} $$
yields
$$\Pr\lbp \lV\frac{S^\intercal U^\intercal \lp n\hat{z} - n\tilde{z}\rp}{m}\rV_\infty \geq t \rbp = \Pr\lbp \lV\frac{S^\intercal U^\intercal \lp \hat{z} - \tilde{z}\rp}{m}\rV_\infty \geq t/n \rbp \leq \frac{\delta}{2}.$$

\end{document}